%% file: template.tex
\documentclass{article}
\usepackage{arxiv}
\usepackage[utf8]{inputenc}
\usepackage[T1]{fontenc}
\usepackage{hyperref, url}
\usepackage{booktabs, array}
\usepackage{amsfonts, amsmath, amssymb, amsthm}
\usepackage{latexsym}
\usepackage{nicefrac}
\usepackage{microtype}
\usepackage{graphicx, multirow}
\usepackage[table]{xcolor}
\definecolor{green}{rgb}{0,.5,0}
\definecolor{magenta}{rgb}{.75,0,.75}
\usepackage[labelfont=bf,font=small]{caption}
\usepackage{float}
\newcommand\scale{1}
\title{Theoretical analysis and experimental validation of volume bias of soft Dice optimized segmentation maps in the context of inherent uncertainty}
\renewcommand{\undertitle}{This is a Medical Image Analysis 2021 preprint}
\renewcommand{\headeright}{MedIA 2021}
\renewcommand{\shorttitle}{Volume bias after soft Dice optimization in the context of inherent uncertainty}
\author{
    \href{https://orcid.org/0000-0001-7206-2671}{\includegraphics[scale=0.06]{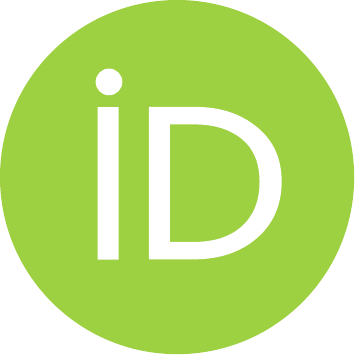}\hspace{1mm}Jeroen Bertels\thanks{Jeroen Bertels and David Robben contributed equally to this work.}}\\
	Processing Speech and Images\\
	Department of Electrical Engineering\\
	KU Leuven, Belgium\\
	\texttt{jeroen.bertels@kuleuven.be}\\
	\And
	David Robben\footnotemark[1]\\
	Processing Speech and Images\\
	Department of Electrical Engineering\\
	KU Leuven, Belgium\\
	\texttt{david.robben@kuleuven.be}\\
	\And
	Dirk Vandermeulen\\
	Processing Speech and Images\\
	Department of Electrical Engineering\\
	KU Leuven, Belgium\\
	\texttt{dirk.vandermeulen@kuleuven.be}\\
	\And
	Paul Suetens\\
	Processing Speech and Images\\
	Department of Electrical Engineering\\
	KU Leuven, Belgium\\
	\texttt{paul.suetens@kuleuven.be}\\
}
\begin{document}
\date{}  
\maketitle
\begin{abstract}
The clinical interest is often to measure the volume of a structure, which is typically derived from a segmentation. In order to evaluate and compare segmentation methods, the similarity between a segmentation and a predefined ground truth is measured using popular discrete metrics, such as the Dice score. Recent segmentation methods use a differentiable surrogate metric, such as soft Dice, as part of the loss function during the learning phase. In this work, we first briefly describe how to derive volume estimates from a segmentation that is, potentially, inherently uncertain or ambiguous. This is followed by a theoretical analysis and an experimental validation linking the inherent uncertainty to common loss functions for training CNNs, namely cross-entropy and soft Dice. We find that, even though soft Dice optimization leads to an improved performance with respect to the Dice score and other measures, it may introduce a volume bias for tasks with high inherent uncertainty. These findings indicate some of the method's clinical limitations and suggest doing a closer ad-hoc volume analysis with an optional re-calibration step.
\end{abstract}
\section{Introduction}\label{sec:introduction}
Automatic segmentation of structures is a fundamental task in medical image analysis. Most automatic segmentation methods learn a single mapping from image to discrete label map~\cite{Kamnitsas2017,Ronneberger2015}. Segmentations either serve as an intermediate step in a more elaborate pipeline or as an end goal by itself. The clinical interest often mainly lies in the volume of segmented structures, such as the volume of a tumor or the volume of a stroke lesion~\cite{Goyal2016}.\\
This paper investigates the estimation of volumes from segmentation maps, more in particular in the context of \textit{inherent uncertainty}. At first sight, inherent uncertainty can be related to observer variability due to image noise, artifacts, limited contrast, hand-eye coordination, fatigue and laziness. But also, inherent uncertainty can be related to the ambiguity of the segmentation task itself. An example of the latter is the prediction of the final infarction (i.e. dead brain tissue), after ischemic stroke treatment (e.g. mechanical thrombectomy to remove a thrombus, thereby allowing brain tissue to get reperfused), from acute imaging information (e.g. CT perfusion)~\cite{Pinto2018, Robben2020}. While a patient's physiology may be fully captured by imaging, without the essential details about the treatment this task remains inherently uncertain. Underperfused brain tissue is dying over time. Depending on how soon the intervention takes place and how much reperfusion will be obtained, the extent of the final infarction will be different. It is clear that this ambiguity relates to the task itself, thus the information available, rather than to the variability observed across manual annotations.\\
In fact, in fields like machine learning and statistics, these sources of inherent uncertainty are typically referred to as \textit{aleatoric uncertainty}~\cite{Kiureghian2009,Hullermeier2021}. In a way these are inherently related to the task description and often acknowledged to result in a non-reducible part of the prediction error. The \textit{epistemic uncertainty} is  considered to result in a reducible part of the prediction error. In a supervised learning setting, the aleatoric uncertainty captures the non-deterministic input-output dependency, while the epistemic uncertainty is roughly an aggregation of both \textit{model} and \textit{approximation uncertainty}. Model uncertainty exists since one can not always guarantee that the hypothesis set, defined by the choice of the model, includes the right model to be fit with respect to the task. Approximation uncertainty is related to the amount of data, loss function, optimizer and other learning parameters, and in essence defined by the discrepancy between the hypothesis produced by the learning algorithm and the one that minimizes the empirical risk. It is further interesting to note that there is no strict separation between aleatoric and epistemic uncertainties and they may merge into one another.\\
In the medical field, the integration of the concept of inherent uncertainty often boils down to making correct confidence predictions alongside the usual outputs. This way, especially with convolutional neural networks (CNNs), good confidence measures would help to establish trustworthiness with the end user. However, recent research by~\cite{Guo2017} revealed that the predictions of modern CNNs are often no longer well-calibrated. They compared various post-processing calibration methods and found that so-called temperature scaling, based on a validation set, could effectively lower the calibration errors on an independent test set. The overall concept of inherent uncertainty or ambiguity of the image segmentation process was recently also addressed by~\cite{Baumgartner2019}, specifically trying to model the conditional probability distribution of the segmentation explicitly given an input image. They focused on the variability that is introduced by different experts, resulting in manual segmentations of different styles.  To a large extent, their research was built on the idea of~\cite{Kohl2018} where similar issues were highlighted as \textit{inherent ambiguities}. Both contributions proposed frameworks using the conditional variational autoencoder~\cite{Kingma2013} in combination with U-Net~\cite{Ronneberger2015}. One focus was on the ability of the method to produce diverse but realistic segmentations, which was studied before in~\cite{Rupprecht2017,Lakshminarayanan2017}. A second focus was to model the conditional distribution correctly (i.e. voxel-level accuracies corresponding to the voxel-level predicted uncertainties), which was studied before in~\cite{Kendall2017,Kendall2019}.\\
Returning to the segmentation task, and the associated estimation of volumes, let us have a look at how such a task is defined and how this has influenced the learning algorithms used during training. To evaluate and compare the quality of a segmentation, the similarity between the true segmentation (i.e. the segmentation derived from an expert's delineation of the structure) and the predicted segmentation must be measured. Common metrics used for this purpose are overlap measures (e.g. Dice score, Jaccard index) and surface distances (e.g. Haussdorf distance, average surface distance)~\cite{Kamnitsas2017, Menze2015}. The focus on Dice score in training CNNs for segmentation led to the definition of a differentiable surrogate loss, the so-called soft Dice~\cite{Drozdzal2016,Milletari2016,Sudre2017}. A large group of state-of-the-art methods now implement this loss function instead of a cross-entropy-based loss function~\cite{Isensee2018}. More recently, it has been shown, both theoretically and experimentally, that optimizing for soft Dice instead of cross-entropy loss indeed improves the final Dice score~\cite{Bertels2019b,Eelbode2020}. To date, to the best of our knowledge, there is no research that investigates whether this evolution in preferred loss function, in combination with the typical inherent uncertainty present in medical applications, had any influence on the resulting segmentations and the derived volume estimates thereof.\\
In this work, we first briefly introduce how to derive volume estimates from segmentation maps. We then relate the overall inherent uncertainty in a segmentation task to a possible volume bias when using different optimization objectives. More specifically, we investigate the volume biases when using cross-entropy or soft Dice, both theoretically and experimentally. We find that the use of soft Dice leads to a systematic under- or over-estimation of the predicted volume of a structure, which is dependent on the inherent uncertainty that is present in the task. We validated these results experimentally on four medical tasks: two tasks with relatively low inherent uncertainty (the segmentation of third molars on dental radiographs~\cite{DeTobel2017,Merdietio2019} and the segmentation of brain tumors on pre-operative MRI in BRATS 2018~\cite{Menze2015,Bakas2017,Bakas2018}) and two tasks with relatively high inherent uncertainty (the segmentation of post-operative infarction based on pre-operative MRI perfusion in ISLES 2017~\cite{Winzeck2018} and the segmentation of the ischemic core on acute CT perfusion in ISLES 2018). Finally, we find that other segmentation metrics benefit from soft Dice optimization and present a simple re-calibration strategy to generate unbiased estimates.
\paragraph{Extension}
This article is an extended version of~\cite{Bertels2020}. The introduction is expanded with links to recent publications that focus on similar definitions of inherent uncertainty. The theoretical section is revised for better readability and made more comprehensive. The experimental section elaborates on the setup in more detail and we ran all experiments with U-Net L (see its definition under Section~\ref{sec:empirical_setup}) to obtain superior segmentation results. We added analyses of relative and absolute volume errors, and included results for other metrics. We also investigated if pre-training with cross-entropy influenced the results. Furthermore, a qualitative analysis is performed, which includes example segmentations, a volume-specific study and a possible re-calibration strategy. The general discussion is extended and an updated conclusion is drawn.
\section{Theoretical analysis}\label{sec:theoretical_analysis}
In this section we first introduce notation and explain why and how a structure's volume needs to be calculated from \textit{soft segmentation maps}. We then investigate what these soft segmentation maps represent in scenarios with non-reducible inherent uncertainty and relate this to volume estimation for binary segmentation tasks.\\
Let us formalize an image into $I$ voxels, each voxel having a true binary class label $l_{i}$ with $i=0 \dots I-1$ ($l_i=0$ for background and $l_i=1$ for foreground), forming the true class label map $l\in \{0,1\}^{I}$. In a similar way we define the input image $x\in \mathbb{R}^{I}$ and the predicted class label map $\tilde{l}\in \{0,1\}^{I}$. In the setting of supervised and gradient-based training of CNNs we are performing empirical risk minimization~\cite{Goodfellow2016}. Image segmentation is generally transformed into a voxel-wise classification task, where each $l_i$ is related to a centered image patch $x^c_i$ extracted around voxel $i$ from $x$. While early methods implemented this voxel-wise classification naively~\cite{Ciresan2012}, state-of-the-art segmentation methods use it implicitly in fully-convolutional setups~\cite{Long2015}.\\
Assume the CNN, with a certain topology, is parametrized by $\theta \in \Theta$ and represents the functions $\mathcal{H}=\{\mathfrak{h}_{\theta}\}^{|\Theta|}$. Further assume we have access to the entire joint probability distribution $P(x^c,l)$ at both training and testing time. The CNN uses $x^c_i$ to make the prediction $\tilde{y}_i=\mathfrak{h}_{\theta}(x^c_i)$ for $l_i$. For these conditions, the general risk minimization principle is applicable and states that in order to optimize the performance for a certain non-negative and real-valued loss $\mathcal{L}$ at test time, we can optimize the same loss during the learning phase~\cite{Vapnik1995}. The risk $\mathcal{R}_{\mathcal{L}}(\mathfrak{h}_{\theta})$ associated with the loss $\mathcal{L}$ and parametrization $\theta$ of the CNN, without regularization, is defined as the expectation of the loss function:
\begin{equation}
    \mathcal{R}_{\mathcal{L}}(\mathfrak{h}_{\boldsymbol{\theta}}) = \mathbb{E} \left[\mathcal{L}\left(\mathfrak{h}_{\boldsymbol{\theta}}(x^c), l\right)\right].
    \label{eq:risk}
\end{equation}\\
Note that the type of output $\tilde{y}$ of the CNN  is typically different from the predicted class label map $\tilde{l}$. This is due to the preferential use of consistent surrogate loss functions with respect to the given discrete losses (i.e. minimizers of the expected surrogate loss also minimize the expected error~\cite{Lapin2018}). In case of binary image classification, we often want to minimize the expected 0/1 error (i.e. accuracy). For this purpose, minimizing the negative log-likelihood has been ubiquitous in terms of risk minimization of CNNs. Its translation into the voxel-wise cross-entropy loss ($\mathcal{CE}$) for binary image segmentation of image $x$ is then given by:
\begin{align}
    \mathcal{CE}(l,\tilde{y}) &= \sum_{i=0}^{I-1}\mathcal{CE}(l_{i},\tilde{y}_{i}) \nonumber\\
    &= -\sum_{i=0}^{I-1}\left[l_{i}\log\tilde{y}_{i}+(1-l_{i})\log(1-\tilde{y}_{i})\right].
    \label{eq:ce}
\end{align}
The activation of the CNN's final layer is typically transformed using the sigmoid function, resulting in a continuous output value $\tilde{y_i} \in [0, 1]$. This will result in a soft segmentation map $\tilde{y}$, from which a predicted label map $\tilde{l}$ can be obtained via:
\begin{equation}
    \tilde{l_i}=
    \begin{cases}
        1, & \text{if } \tilde{y}_i \geq 0.5\\
        0, & \text{if } \tilde{y}_i < 0.5
    \end{cases}.
    \label{eq:thresholding}
\end{equation}\\
To better evaluate the segmentation quality, compared to voxel-wise accuracy, overlap measures are frequently used. In the context of medical image analysis,~\cite{Zijdenbos1994} were among the first to use the Dice score ($\mathcal{D}$)~\cite{Dice1945} for evaluating the quality of white matter lesion segmentation:
\begin{equation}
    \mathcal{D}(l, \tilde{l}) = \frac{2\sum_{i=0}^{I-1}(l_{i}\tilde{l}_{i})}{\sum_{i=0}^{I-1}l_{i} + \sum_{i=0}^{I-1}\tilde{l}_{i}}.
    \label{eq:dice}
\end{equation}
More recently, the soft Dice loss ($\mathcal{SD}$)~\cite{Drozdzal2016,Milletari2016,Sudre2017} has been used in the optimization of CNNs to directly optimize the Dice score at test time~\cite{Bertels2019a}. Note that we can define $\mathcal{SD}$ as one minus the soft Dice score for minimization. Rewriting Eq.~\ref{eq:dice} to its non-negative and real-valued surrogate loss function as in~\cite{Drozdzal2016}, with the $L_1$ norm an acceptable choice~\cite{Eelbode2020}, we get for image $x$:
\begin{equation}
    \mathcal{SD}(l,\tilde{y}) = 1-\frac{2\sum_{i=0}^{I-1}(l_{i}\tilde{y}_{i})}{\sum_{i=0}^{I-1}l_{i} + \sum_{i=0}^{I-1}\tilde{y}_{i}}.
    \label{eq:softdice}
\end{equation}\\
In the following sections we will analyze the influence of the inherent uncertain character of the segmentation task to volume estimations. In binary segmentation tasks, the soft output maps $\tilde{y}$ of the CNN are typically binarized using Eq.~\ref{eq:thresholding}. The volumes $\mathcal{V}(l)$ of the true structure and $\mathcal{V}(\tilde{l})$ of the predicted structure, with $v$ the volume of a single voxel, are then given by:
\begin{align}
    \mathcal{V}(l) &= v\sum_{i=0}^{I-1}l_{i} \label{eq:true_volume},\\
    \mathcal{V}(\tilde{l}) &= v\sum_{i=0}^{I-1}\tilde{l}_{i}.
\end{align}
\subsection{Inherent uncertainty}\label{seq:inherent_uncertainty}
As mentioned in Section~\ref{sec:introduction}, segmentation of medical images is an inherently ambiguous task. Indeed, images might be noisy, lack contrast, contain artifacts or just lack complete information. Even at the level of the ground truth segmentation, uncertainty is introduced due to intra- and inter-observer variability. In what follows we will investigate what happens with the estimated volume of segmented structures in an image under the assumption of having perfect segmentation algorithms (i.e. the hypothesis set contains the model that fits the task and the prediction is the one that minimizes the empirical risk), and thus with only the formerly described non-reducible part of the inherent uncertainty remaining.\\
Assume we have for the image $x$ a true label map $l$ that we model now as a discrete random field. The label $l_i$ of every voxel $i$ is inherently uncertain, with $y_i \in [0, 1]$ reflecting the probability of belonging to the foreground structure. In this case the calculation of the true volume from Eq.~\ref{eq:true_volume} boils down to working out the expectation:
\begin{equation}
    \mathbb{E}[\mathcal{V}(l)] = v\mathbb{E} [\sum_{i=0}^{I-1}l_{i}] = v \sum_{i=0}^{I-1}\mathbb{E}[l_{i}] = v \sum_{i=0}^{I-1}y_{i} = \mathcal{V}(y).
    \label{eq:true_volume_expectation}
\end{equation}
Note that we do not write out a similar equality for the expectation over the predicted volume $\mathbb{E}[\mathcal{V}(\tilde{l})]$. This would only be correct if the soft output map $\tilde{y}$ indeed represents a probability distribution with $P(\tilde{l}_i=1|x_i^c,\theta)=\tilde{y}_i$. The valid use of $\mathcal{V}(\tilde{y})$ to estimate $\mathbb{E}[\mathcal{V}(l)]$ will thus depend on the relationship between $\tilde{y}$ and $y$. This will be investigated next in more detail for both $\mathcal{CE}$ and $\mathcal{SD}$ optimization.
\paragraph{Effect of $\mathcal{CE}$ optimization} 
If we want to analyze for $\mathcal{CE}$ the predictions $\tilde{y}$ that minimize the risk $\mathcal{R}_{\mathcal{CE}}(\mathfrak{h}_{\boldsymbol{\theta}})$ we need to work out:
\begin{align}
    \arg\min_{\tilde{y}}\mathcal{R}_{\mathcal{CE}}(\mathfrak{h}_{\theta}) = &\arg\min_{\tilde{y}}\mathbb{E} [\mathcal{CE}(l,\tilde{y})] \nonumber\\
    = &\arg\min_{\tilde{y}_i}\sum_{i=0}^{I-1} \mathcal{CE}(y_i,\tilde{y}_i).\label{eq:voxel-wise}
\end{align}
We need to find for each voxel $i$ with true foreground probability $y_i$ the soft prediction $\tilde{y}_i \in [0,1]$ independently, hence:
\begin{align}
    &\arg\min_{\tilde{y}_{i}}\mathcal{CE}(y_{i},\tilde{y}_i) \nonumber\\
    = &\arg\min_{\tilde{y}_{i}}[-y_{i}\log\tilde{y}_i-(1-y_{i})\log(1-\tilde{y}_{i})].
    \label{eq:cemin}
\end{align}
This function is continuous and its first derivative monotonically increasing in the interval $]0, 1[$. First order conditions with respect to $\tilde{y}_{i}$ give the optimal value for the predicted uncertainty: $\tilde{y}_{i}=y_{i}$. With the predicted soft output map $\tilde{y}$ representing the true inherent uncertainty after $\mathcal{CE}$ optimization, $\mathcal{V}(\tilde{y})$ becomes an unbiased estimator for $\mathbb{E}[\mathcal{V}(l)]$ .
\paragraph{Effect of $\mathcal{SD}$ optimization} 
If we want to analyze for $\mathcal{SD}$ the predictions $\tilde{y}$ that minimize the risk $\mathcal{R}_{\mathcal{SD}}(\mathfrak{h}_{\boldsymbol{\theta}})$ we need to work out:
\begin{equation}
    \arg\min_{\tilde{y}}\mathcal{R}_{\mathcal{SD}}(\mathfrak{h}_{\theta}) = \arg\min_{\tilde{y}}\mathbb{E} [\mathcal{SD}(l,\tilde{y})].\label{eq:R_SD_}
\end{equation}
We need to find for each voxel $i$ with true foreground probability $y_i$ the soft prediction $\tilde{y}_i \in [0,1]$, hence:
\begin{align}
    &\arg\min_{\tilde{y}_{i}}\mathbb{E}\left[1-\frac{2\sum_{i=0}^{I-1}(l_{i}\tilde{y}_{i})}{\sum_{i=0}^{I-1}l_{i} + \sum_{i=0}^{I-1}\tilde{y}_{i}}\right].
\end{align}
This minimization is more complex and we analyze its behavior by inspecting the values of $\mathcal{SD}$ numerically for some specific scenarios that are visualized in Figure~\ref{fig:inherent_uncertain_regions_with_volumes}.
\input{figures/inherent_uncertain_regions_with_volumes.tex}
\subsection{Numerical simulation for $\mathcal{SD}$}
Assume the labels of each voxel are conditionally independent, or that we can simplify the segmentation of an image into $J$ independent regions with true foreground probability $p_{j}$ and volumes $s_{j}=vn_{j}$, and with $n_{j}$ the number of voxels belonging to region $j=0 \dots J-1$ (for $n_{j}=1$ and  $J=I$, each voxel is an independent region), we get:
\begin{equation}
    \mathcal{V}(l) = v \sum_{j=0}^{J-1} (n_{j}p_{j}) = \sum_{j=0}^{J-1} (s_{j}p_{j}).
    \label{eq:true_volume_regions}
\end{equation}
This situation is depicted in Figure~\ref{fig:inherent_uncertain_regions_with_volumes} for an increasing number (left to right) of inherently uncertain regions. The white area is $100 \%$ foreground, the black region is $100 \%$ background, the grey areas (separated by the dashed radial lines) are conditionally independent regions and inherently uncertain to belong to the foreground structure.\\
Referring to Eq.~\ref{eq:R_SD_} we now need to find for each independent region $j$ with true foreground probability $p_j$ the soft prediction $\tilde{p}_j \in [0,1]$, hence:
\begin{align}
    &\arg\min_{\tilde{p}}\mathbb{E} [\mathcal{SD}(l,\tilde{p}) ]\nonumber\\
    = &\arg\min_{\tilde{p}_{j}}\mathbb{E}\left[1-\frac{2\sum_{j=0}^{J-1}(l_{j}s_j\tilde{p}_{j})}{\sum_{j=0}^{J-1}l_{j} + \sum_{j=0}^{J-1}(s_j\tilde{p}_{j})}\right]. \label{eq:R_SD}
\end{align}
In order to examine the influence of the number of regions and their sizes on the solutions of Eq.~\ref{eq:R_SD_}, we consider the specific scenarios with $K=1$, $4$ or $16$ independent regions $\beta_k$ ($k=0...K-1$) with inherent uncertainty. For each scenario we vary the inherent uncertainty by varying the true foreground probability $p_{\beta_k}$ and the total uncertain volume $\sum_{k=0}^{K-1}s_{\beta_k}$ (colors), and analyze their effect in terms of under- or over-estimation of the predicted foreground probability $\tilde{p}_{\beta_k}$, and thus indirectly the volume bias $\mathbb{E}[\Delta \mathcal{V}]$.
\paragraph{Single region with inherent uncertainty ($K=1$)}
\input{figures/all_regions_plot.tex}
\input{figures/p_bias_plot.tex}
Imagine the segmentation of the left image in Figure~\ref{fig:inherent_uncertain_regions_with_volumes}. This image has a background region $\alpha$ (black), a single region $\beta_0=\beta$ that is inherently uncertain (gray), and a foreground region $\gamma$ (white). The region $\alpha$ is certainly not part of the structure and thus $p_{\alpha}=0$. The region $\beta$ belongs to the structure with true foreground probability $p_{\beta} \in [0,1]$. The region $\gamma$ is certainly part of the structure with $p_{\gamma}=1$. Assuming a perfect algorithm, the optimal predictions under the empirical risk (Eq.~\ref{eq:R_SD}) for a single region with inherent uncertainty are given by:
\begin{equation}
    \arg\max_{\tilde{p}_{\alpha},\tilde{p}_{\beta},\tilde{p}_{\gamma}}\mathbb{E}\left[\frac{2(s_{\beta}l_{\beta}\tilde{p}_{\beta}+s_{\gamma}\tilde{p}_{\gamma})}{s_{\alpha}\tilde{p}_{\alpha} + s_{\beta}\tilde{p}_{\beta} + s_{\gamma}\tilde{p}_{\gamma} + s_{\beta}l_{\beta} + s_{\gamma}}\right].
\end{equation}
It is trivial to show that $\tilde{p}_{\alpha}=0=p_{\alpha}$ and $\tilde{p}_{\gamma}=1=p_{\gamma}$ are solutions for this equation. The behavior of $\mathbb{E}[\mathcal{SD}(l,\tilde{p})]$ as a function of $\tilde{p}_{\beta}$ for $K=1$ is presented in the left column of Figure~\ref{fig:all_regions_plot}. In each plot, a higher line opacity corresponds to a higher true foreground probability $p_\beta=\{0, 0.25, 0.5, 0.75, 1\}$ and different colors point to a different total volume of the inherently uncertain region. Denote the volumes of region $\alpha$, $\beta$ and $\gamma$ by  $s_{\alpha}$, $s_{\beta}$ and $s_{\gamma}$, with $s_{\alpha}=100$ and $s_{\gamma}=1$. We can denote $\mu=s_{\beta}/s_{\gamma}=s_{\beta}$ as the volume ratio of the total uncertain to certain part of the structure. In Figure~\ref{fig:inherent_uncertain_regions_with_volumes}, the uncertain region $\beta$ starts at the border of the white foreground region $\gamma$ and extends to the blue ($\mu=0.25$), black ($\mu=1$) or red ($\mu=4$) circle, where the background region $\alpha$ starts, corresponding to the different rows in Figure~\ref{fig:all_regions_plot}. In the left plot of Figure~\ref{fig:p_bias_plot} the error on the predicted foreground probability $\tilde{p}_\beta-p_\beta$ (at optimal $\tilde{p}_\beta$) is plotted as a function of the inherent uncertainty in terms of the true foreground probability $p_\beta$.\\
As a first observation, in the left column in Figure~\ref{fig:all_regions_plot} we note that the local minimum of the  $\mathbb{E}[\mathcal{SD}]$ loss function is either at $\tilde{p}_\beta=0$ or $\tilde{p}_\beta=1$. As such, we can reconfirm that only for $p_{\beta}=\{0, 1\}$ the predicted foreground probability $\tilde{p}_{\beta}$ is exact. The location of the local minimum of $\tilde{p}_{\beta}$ in $[0, 1]$ switches from 0 to 1 when $p_{\beta}=0.5$. This supports the second observation in Figure~\ref{fig:p_bias_plot}: for $p_{\beta}$ smaller or larger than 0.5, respectively under- or over-estimation will occur. This results in both an error on the predicted foreground probability $\tilde{p}_\beta$ as well as a bias on the derived volume. The resulting volume bias will be maximal when the true foreground probability is $p_{\beta}=0.5$ (maximal uncertainty) and decreases towards the points of complete certainty, being always 0 or 1. The effect of the volume ratio $\mu$ is two-fold. First, with $\mu$ increasing, the optimal  $\mathbb{E}[\mathcal{SD}]$ loss value increases. Second, although the error on the estimated foreground probability is not influenced by $\mu$ (all lines are on top of each other in Figure~\ref{fig:p_bias_plot}), the volume bias increases (the same error on the foreground probability gets multiplied with a larger volume).
\paragraph{Multiple regions with inherent uncertainty ($K=\{4,16\}$)}
In a realistic scenario we can expect multiple independent regions. Think of the uncertainty introduced due to observer variability at the borders of a tumor. In a way the delineation at one side of the tumor is independent of the delineation at the other side. Similarly, we can simulate the segmentation of a structure with $K=\{4, 16\}$ independent regions with inherent uncertainty (middle and right images in Figure~\ref{fig:inherent_uncertain_regions_with_volumes}). Region $\beta$ is now further subdivided into 4 or 16 equally large and independent sub-regions $\beta_{k}$ with equal true foreground probability $p_{\beta_{k}}=p_{\beta}$ and volume ratio $\mu_{\beta_{k}}$  and a total volume ratio $K*\mu_{\beta_{k}}=\mu_{\beta}$  to keep the total uncertain volume the same as for $K=1$. We limit the analysis to a qualitative observation of the middle and right columns of Figure~\ref{fig:all_regions_plot} and Figure~\ref{fig:p_bias_plot}, with $K=4$ and $K=16$, respectively. As a first observation, in Figure~\ref{fig:all_regions_plot} we notice that the local minimum of $\mathbb{E}[\mathcal{SD}]$ is still at $\tilde{p}_\beta=0$ or $\tilde{p}_\beta=1$. However, we further notice that the true foreground probability $p_{\beta}$ for which under- or over-estimation will occur decreases compared to a single region of uncertainty ($K=1$; previous paragraph; left columns). This is visible in each row if we compare lines with the same opacity from different columns. Note for example for $\mu=4$ (bottom row/red color) that the middle line flipped and now has a minimum at $\tilde{p}_\beta=1$. Second, from Figure~\ref{fig:p_bias_plot} we see that this effect is more pronounced when $K$ increases (note that we see for each color a larger shift from the mid-line to the left in the right plot compared to the middle plot). Third, this effect is also proportional with $\mu$ and the maximal error on the predicted foreground probability becomes higher (in Figure~\ref{fig:p_bias_plot}, for $K > 1$ we see that the shift to the left is proportional with $\mu$). Fourth, when the number of regions increases there is a trend towards volume over-estimation (with a larger maximal error) due to a larger range of $p_\beta$ corresponding to volume over-estimation.
\subsection{Summary of theoretical analysis}\label{sec:theoretical_summary}
From the theoretical and numerical analysis we conclude that $\mathcal{CE}$ and $\mathcal{SD}$ optimization will lead to different volume estimators. Because $\mathcal{CE}$ is able to produce correct foreground probabilities, its soft output segmentations $\tilde{y}$ can be used directly for summation to produce $V(\tilde{y})$, an unbiased estimator for $\mathbb{E}[V(l)]$. Note that in this case $V(\tilde{y})$ can be different from $V(\tilde{l})$ due to inherent uncertainty. However, $\mathcal{SD}$ optimization prefers binary output segmentations, with no difference between $V(\tilde{y})$ and $V(\tilde{l})$, thus producing biased volume estimates when there is inherent uncertainty in the task. More specifically, we expect for $\mathcal{SD}$ a trend towards over-estimation due to the larger range of inherent uncertainties that will lead to an over-estimation of the foreground probability.
\section{Experimental validation}
In this section we will investigate whether the aforementioned characteristics can be observed under real circumstances. In a practical scenario, the complete joint probability distribution $P(x^c,l)$ is not known and can only be estimated based on the training set samples. Hence, the risk $\mathcal{R}_{\mathcal{L}}$ (Eq.~\ref{eq:risk}) becomes empirical, where the expectation of the loss function becomes the mean of the losses across the training set. Furthermore, the loss $\mathcal{L}$ absorbs the explicit (e.g. weight decay, L2) or implicit (e.g. early stopping, dropout) regularization, which is often present in some aspect of the optimization of CNNs. Finally, the classifier is no longer perfect and as a result, in addition to the non-reducible inherent uncertainty in the task, we now have added reducible inherent uncertainty introduced by the classifier itself.
\subsection{Experimental setup}\label{sec:empirical_setup}
To investigate how these factors impact our theoretical findings, we train four models with increasing complexity. The first model performs a logistic regression on the input features. The other three models are U-Net-like~\cite{Ronneberger2015} with varying complexity. The models will be optimized using five-fold cross-validation on four different medical tasks and with respect to both $\mathcal{CE}$ and $\mathcal{SD}$ objectives. The expected levels of non-reducible inherent uncertainty in the tasks vary and are explained below.\\\\
In the first two tasks, the ground truth is manually delineated on the input images. Hence the only source of uncertainty is in the delineation process. In the third and fourth task, the ground truth is delineated on images that are not used as input. As such, there is both uncertainty due to the delineation and due to the difference in modality.
Tasks with a relatively low level of inherent uncertainty, due to the manual delineations being defined on the same inputs, are:
\begin{itemize}
    \item MO17: \textit{lower-left third molar segmentation on panoramic dental radiographs}~\cite{DeTobel2017}. The manual delineation guidelines were defined on the same inputs. In theory, all information is available, however in practice, inherent uncertainty will be introduced due to observer variability (e.g. due to image quality or laziness).
    \item BR18: \textit{segmentation of whole brain tumors on pre-operative MRI} in BRATS 2018~\cite{Menze2015,Bakas2017,Bakas2018}. Similar reasoning to MO17. The individual tumor structures available in BRATS (i.e. nectrotic with non-enhancing tumor core, peritumoral edema, enhancing core, everything else) might result in more inherent uncertainty due to lower image contrast and more experience required.
\end{itemize}
Tasks with a relatively high level of inherent uncertainty, due to the manual delineations being defined on different inputs, are:
\begin{itemize}
    \item IS17: \textit{segmentation of post-operative final infarctions based on pre-operative MRI data} in ISLES 2017~\cite{Winzeck2018}. While the definition of the manual segmentation is based on registered post-operative CT or MRI data, the inputs are pre-operative parameter maps. This results in a large amount of ambiguity in the task description when only taking into account imaging data and leaving out treatment information.
    \item IS18: \textit{segmentation of ischemic cores based on acute CT data} in ISLES 2018. While the definition of the manual segmentation is based on registered pre-operative MRI data, the inputs are pre-operative perfusion parameter maps derived form a CT perfusion. It is well-known that CT and MRI provide different, rather complementary, information to retrieve the true ischemic core from acute imaging~\cite{VonKummer2017}. Moreover, the lower spatial resolution of a CT perfusion further increases the level of inherent uncertainty.
\end{itemize}
Next, we give more details on the processing of these datasets, network architectures, training and statistical testing. This will be followed by a presentation of the quantitative results, zooming in on the predicted volume errors $\Delta \mathcal{V}$, and a qualitative inspection of the segmentation outputs.
\paragraph{Datasets}
We formulate a binary segmentation task for each dataset having one (multi-modal) input, and giving one binary segmentation map as output. For BR18 we limit the task to whole tumor segmentation. For all three 3D public benchmarks (i.e. BR18, IS18, IS17) we use all of the provided images, except for IS17 and IS18 where the native perfusion series were omitted. As a result, for BR18 we concatenate the MRI sequences \{T\textsubscript{1}, contrast-enhanced T\textsubscript{1}, T\textsubscript{2}, FLAIR\}, for IS17 we concatenate the parameter maps \{ADC, T\textsubscript{max}, MTT, TTP, rCBF, rCBV\} and for IS18 we concatenate the CT together with the parameter maps \{T\textsubscript{max}, MTT, CBF, CBV\}. These inputs are spatially resampled to an isotropic voxel-size of 2 mm. We use linear interpolation when resampling the label maps, keeping soft values for training. In the 2D dataset MO17, we first extract a 448x448 ROI around the geometrical center of the lower-left third molar from the panoramic dental radiograph and further downsample the ROI by a factor of two. The output is the segmentation of the third molar, as provided by the experts. All images are normalized according to the dataset's mean and standard deviation. Some characteristics of the resulting datasets used for the experimental analysis are:
\begin{itemize}
    \item MO17: 400 cases of size 224x224;
    \item BR18: 285 cases of size 120x120x78;
    \item IS17: 43 cases in size range 115x115x62-120x120x78;
    \item IS18: 94 cases in size range 102x102x10-133x133x80.
\end{itemize}
\paragraph{Network architecture}
\input{figures/networks.tex}
In Figure~\ref{fig:networks} the network architectures for all four models are illustrated. The architecture of the LR model is the part inside the top-left dark gray rectangle and uses the inputs directly for classification, thus performing logistic regression on the input features. For our three U-Net-like models we start from the successful No New-Net implementation during the BRATS 2018 challenge~\cite{Isensee2018} and modify this to alter the complexity, respectively U-Net S, U-Net M and U-Net L for increasing complexity. The specific architecture for U-Net S is within the light gray rectangle with $n=10$ feature maps in the first layer. For the two more complex networks this extends up until the dashed line, with $n=10$ and $n=20$ for U-Net M and U-Net L, respectively. The major modifications compared to~\cite{Isensee2018} are that we used average pooling and that we did not use instance normalization~\cite{Ulyanov2016} for U-Net S and U-Net M. The U-Net L model closely resembles No New-Net with only a slightly lower number of parameters to enable full-image processing at an isotropic voxel-size of 2 mm. We further opted for valid trilinear upsampling, thereby preserving spatial alignment, and which explains the discrepancy between input and output sizes of the networks.
\paragraph{Training}
All of the input images are augmented intensively during training by adding Gaussian noise, performing small random translations and in-plane rotations, and by allowing lateral flips [in an additional experiment this process was confirmed not to introduce any volume bias]. The inputs are presented as central image crops of 162x162x108 (in MO17 243x243) to the networks, both during training and testing, with zero padding. As such, the outputs are of size 136x136x82 (in MO17 217x217) and in a similar way these are cropped to the size of the ground truth. We train all models with respect to the $\mathcal{CE}$ or $\mathcal{SD}$ objective with the ADAM optimizer~\cite{Kingma2014a}. Any explicit regularization was avoided, except for U-Net L where a small L2 regularization of $10^{-5}$ was helpful to obtain robust convergence. The initial learning rate was set at $10^{-3}$ (for LR model at 1) and lowered by a factor of five when the validation loss did not improve over the last 75 epochs (for BR18 150 epochs). Training was stopped when no improvement occurred over the last 150 epochs (for BR18 300 epochs) and the final model was chosen according to the optimization objective over the validation set.
\paragraph{Statistical testing}
Non-parametric bootstrapping was used to assess inferiority and superiority between groups (e.g. between the group of individual Dice scores on the aggregated validation sets coming from U-Net L after $\mathcal{CE}$ or $\mathcal{SD}$ optimization) at a significance level of $p < 0.05$. We therefore sampled 10000 times an equally-sized group from the pair-wise differences with replacement. We further note that in this work cross-validation was used, at random, without the use of additional independent test sets.
\subsection{Experimental results}
First, the overall quantitative results that are directly related with the main conclusions from the theoretical analysis will be presented, followed by a quantitative inspection of other measures. For U-Net L, the quantitative results after $\mathcal{CE}$ pre-training will also be investigated. Second, we present a qualitative inspection through the display of the segmentation maps, look into the relation between volume and volume error and present a simple re-calibration strategy.
\subsubsection{Quantitative evaluation}
\paragraph{Volume bias}
Table~\ref{tab:table_volumetric_bias} shows the volume bias $\mathbb{E}[\Delta\mathcal{V}]$ (i.e. mean volume error over the validation set) for the four datasets for the four models optimized with respect to $\mathcal{CE}$ or $\mathcal{SD}$. The volume bias is given for the volumes calculated with the predicted soft or thresholded segmentations, respectively $\mathbb{E}[\Delta\mathcal{V}(\tilde{y},l)]$ and $\mathbb{E}[\Delta\mathcal{V}(\tilde{l},l)]$. As a first observation, and as expected from Section~\ref{sec:theoretical_summary}, we note that for $\mathcal{CE}$ optimized models $\mathbb{E}[\Delta\mathcal{V}(\tilde{y},l)]$ is different from $\mathbb{E}[\Delta\mathcal{V}(\tilde{l},l)]$, with the former having smaller bias. For $\mathcal{SD}$ optimized models this difference is negligible. As a second observation, also as expected from Section~\ref{sec:theoretical_summary}, the optimization with respect to $\mathcal{SD}$ leads to over-estimation. This is highly significant ($p < 0.001$) for simple models and remains significant for U-Net M for IS17 and IS18 and for U-Net L for IS18 (and a trend remains for IS17), the two medical tasks having the highest inherent uncertainty. For $\mathcal{CE}$ optimized models this bias is almost absent.\\
The boxplots in Figure~\ref{fig:boxplots} give a view on the distribution of $\Delta\mathcal{V}(\tilde{y},l)$ (left side in each boxplot) and $\Delta\mathcal{V}(\tilde{l},l)$ (right side in each boxplot) for U-Net L. As a first observation, it is clear that $\mathcal{CE}$ optimization (green color) results are generally unbiased with respect to $\Delta\mathcal{V}(\tilde{y},l)$, in contrast to $\Delta\mathcal{V}(\tilde{l},l)$, which might result in a lower spread of the individual data points. This discrepancy is absent for $\mathcal{SD}$ optimization. As a second observation, it is clear that for $\mathcal{SD}$ a volume bias remains for tasks with high inherent uncertainty. However, $\mathcal{SD}$ optimization has a beneficial effect on the spread of the individual volume errors. From the BR18 boxplot it is clear that there is one outlier with strong under-estimation for all settings, which might explain the negative bias seen in Table~\ref{tab:table_volumetric_bias}.
\input{tables/table_volumetric_bias.tex}
\input{figures/boxplots.tex}
\paragraph{Other measures}
Table~\ref{tab:table_other_measures} shows final loss values and the corresponding metrics for U-net M and U-Net L. We also report the relative volume errors 
and their absolute interpretations since comparing absolute volume differences and/or normalized to their manual volumes is of significant practical value (e.g.~\cite{Kuijf2019,Ermis2020}). As a first observation, $\mathcal{CE}$ and $\mathcal{SD}$ optimization leads to the minimization of the corresponding risks (the soft Dice score is given, which is $1-\mathcal{SD}$ and should be maximized). Similarly, the voxel-wise accuracy (i.e. 0/1 metric) and Dice score are metrics that were indeed optimized by their corresponding relaxations. For U-Net M and U-Net L this was often insignificant, however, for LR and U-Net S the results were always significant and in favor of the corresponding relaxations. As a second observation, looking at the relative volume bias $\mathbb{E}[\Delta\mathcal{V}/\mathcal{V}(l)]$, we see that neither $\mathcal{CE}$ optimization nor $\mathcal{SD}$ optimization lead to consistent results. For U-Net M one could still argue a biasing effect of $\mathcal{SD}$ optimization, however, for U-Net L it looks like the relative nature of $\mathcal{SD}$ helps in the optimization of the relative volume bias. A similar observation can be made for the expected relative absolute volume error $\mathbb{E}[|\Delta\mathcal{V}|/\mathcal{V}(l)]$. It looks like $\mathcal{SD}$ optimization works particularly well in more complex models, thereby producing equal or better results in terms of relative bias compared to $\mathcal{CE}$ optimization. The larger spreads that were observed in Figure~\ref{fig:boxplots} are quantitatively confirmed with a larger absolute volume error for $\mathcal{CE}$ compared to $\mathcal{SD}$ optimization.
\input{tables/table_other_measures.tex}
\paragraph{Effect of $\mathcal{CE}$ pre-training}
Table~\ref{tab:table_CE_pre-train} shows the volume bias after initial $\mathcal{CE}$ optimization for U-Net M and U-Net L. For this purpose, we used the $\mathcal{CE}$ optimized models and repeated the training procedure with both $\mathcal{CE}$ and $\mathcal{SD}$. This test is performed given, first, the frequent use of $\mathcal{CE}$ pre-trained networks as an alternative for building ad-hoc networks in combination with transfer learning~\cite{Chen2016,Iglovikov2018} and, second, the often beneficial convergence properties of $\mathcal{CE}$~\cite{Bertels2019a}. We note that the results regarding volume bias are generally reproduced, however, post-$\mathcal{CE}$ optimization suffers from under-estimation in some cases, which again for BR18 might be the result of the one outlier visible in the boxplot in Figure~\ref{fig:boxplots}.
\input{tables/table_CE_pretrain.tex}
\subsubsection{Qualitative inspection}
\paragraph{Segmentation examples}
\input{figures/qualitative_inspection_ce_dice.tex}
In Figure~\ref{fig:qualitative_inspection_ce_dice} example segmentation maps that resulted from models LR, U-Net S, U-Net M and U-Net L are visualized after $\mathcal{CE}$ (top) or $\mathcal{SD}$ (bottom) optimization. As a first observation, looking at the $\mathcal{CE}$ outputs, we see soft maps that evolve from somewhat empty maps to more condensed segmentations with increasing model complexity or decreasing inherent uncertainty in the task description. This is in sharp contrast with the corresponding $\mathcal{SD}$ outputs delivering highly binary-like segmentation maps. As a second observation, looking at the $\mathcal{CE}$ outputs, we can understand why quantitative differences were observed when using $\mathcal{V}(\tilde{y}, l)$ or $\mathcal{V}(\tilde{l}, l)$ as the volume estimate. Due to thresholding at 0.5 under-segmentation of the structure will occur for simpler models, thereby resulting in under-estimation of the volume. A similar but reverse observation can be made for $\mathcal{SD}$ optimization. It is clear that $\mathcal{SD}$ optimization results in over-segmentation, which in turn leads to over-estimation of the volume, as expected. However, the better the models can discriminate foreground from background, the smaller this over-estimation will be.\\
Let us perform the thought experiment on Figure~\ref{fig:qualitative_inspection_ce_dice} of superimposing the $\mathcal{SD}$ delineations (bottom) on the soft segmentation maps resulting from $\mathcal{CE}$ optimization (top). While reducing the level of inherent uncertainty due to increasing model complexity (from left to right) or due to less ambiguity in the task description (bottom two rows to top two rows) we notice that the $\mathcal{CE}$ and $\mathcal{SD}$ delineations converge. It looks like if a predicted $\mathcal{SD}$ delineation is mostly contained within a lower-thresholded $\mathcal{CE}$ delineation and thus $\mathcal{SD}$ delineations extending further into the uncertainty.
\paragraph{Volume-specific analysis}
In Figure~\ref{fig:scale_specific_study_and_recalibration}-top individual volume predictions $\mathcal{V}(\tilde{y})$ or $\mathcal{V}(\tilde{l})$, respectively after $\mathcal{CE}$ or $\mathcal{SD}$ optimization, are shown for U-Net L as a function of the true volume $\mathcal{V}(l)$. As a first observation, we notice for both $\mathcal{CE}$ and $\mathcal{SD}$ optimization a volume-specific bias that over-estimates small volumes and under-estimates large volumes. The transition from under- to over-estimation seems to happen around the average true volume in the dataset. As a second observation, we notice a slightly larger spread on the individual data points and a larger volume-specific bias after $\mathcal{CE}$ optimization compared to $\mathcal{SD}$.
\paragraph{Re-calibration}\label{sec:re-calibration}
In Figure~\ref{fig:scale_specific_study_and_recalibration}-bottom the volume-specific results are shown after re-calibration. In order to correct for the volume-specific bias, and for the volume bias, we analyzed for each fold the volume-specific behavior on the training set [data not shown]. We used least-squares regression to fit for each fold the linear model that best explains the volume-specificity and bias on the training set and use this to correct for on the validation set. The training data showed similar volume-specific biases compared to the biases in the validation data, however, the data was much closer to the unity line. Nonetheless, comparing Figure~\ref{fig:scale_specific_study_and_recalibration}-top with Figure~\ref{fig:scale_specific_study_and_recalibration}-bottom we notice a decrease in volume-specificity of the volume estimates, with data slope trends closer to one. Furthermore, a quantitative inspection confirmed that there was no significant bias $\mathbb{E}[\Delta\mathcal{V}]$ remaining after re-calibration, except on BR18 where after $\mathcal{SD}$ optimization the bias was reduced from -3.90 to -2.34 ml, but remained significantly different from zero. While all volume-specificity slopes were different from zero before re-calibration, after re-calibration no significant slope for the $\mathcal{CE}$ model of IS17 and the $\mathcal{CE}$ and $\mathcal{SD}$ models of IS18 remained.
\input{figures/scale_specific_study_and_recalibration.tex}
\subsubsection{Summary of experimental validation}
From this experimental validation we conclude that, in order to be an unbiased estimator, $\mathcal{CE}$ requires the volumes to be calculated using $\mathcal{V}(\tilde{y})$. Despite being unbiased, the individual data points calculated after $\mathcal{SD}$ optimization have smaller spread and overall $\mathcal{SD}$ optimization may have beneficial properties regarding relative and absolute volume errors. A volume-specific analysis revealed a smaller error after $\mathcal{SD}$ compared to $\mathcal{CE}$ optimization and a potentially useful re-calibration strategy was proposed to correct for bias using the results on the training set. Nonetheless, depending on the remaining inherent uncertainty (related to both the ambiguity in the task description and the model complexity), $\mathcal{SD}$ optimization may indeed result in a volume bias, with or without $\mathcal{CE}$ pre-training.
\section{Discussion}
To the best of our knowledge, this is the first study to relate volume estimation with the optimization settings of a CNN-based methodology. The theoretical findings that were listed under~\ref{sec:theoretical_summary} were confirmed by the experimental validation. On the one hand, $\mathcal{CE}$ optimization estimates the voxel-wise foreground probabilities (Eq.~\ref{eq:cemin}), which indeed resulted in so-called soft segmentation maps, illustrated in Figure~\ref{fig:qualitative_inspection_ce_dice}-top. A closer view on this was recently established by~\cite{Jungo2019}, showing that this holds experimentally across the entire dataset, although for individual data points the predictions might not be well-calibrated, as was similarly observed by~\cite{Guo2017} for recent CNNs in general. On the other hand, we could numerically show that $\mathcal{SD}$ optimization leads to binary maps, which was indeed the case in Figure~\ref{fig:qualitative_inspection_ce_dice}-bottom. For $\mathcal{CE}$ optimization this proceeded into unbiased volume estimates on the dataset level when using $\mathcal{V}(\tilde{y})$ as the predictor (Table~\ref{tab:table_volumetric_bias} and Figure~\ref{fig:boxplots}). For $\mathcal{SD}$ optimization the expected over-segmentation occurred and seemed to remain persistent for tasks with high inherent uncertainty (Table~\ref{tab:table_volumetric_bias} and Figure~\ref{fig:boxplots}). Even under realistic experimental settings the theoretical conclusions after applying the general risk minimization principle remained valid. This also includes the use of $\mathcal{CE}$-based pre-training, which is often observed in practice either via transfer learning~\cite{Chen2016,Iglovikov2018} or to improve the convergence~\cite{Bertels2019b,Eelbode2020}, and which could potentially hinder the solution of $\mathcal{SD}$ being optimal.\\
In the experimental results it was further investigated what effect $\mathcal{CE}$ and $\mathcal{SD}$ optimization had on the relative and absolute volume errors, which are often used in challenges (e.g.~\cite{Kuijf2019}) or serve more practical value (e.g.~\cite{Ermis2020}). It was clear that uncertainty was negatively influencing the performance of $\mathcal{SD}$ optimization more compared to $\mathcal{CE}$ optimization. However, as soon as models get more complex or when tasks carry less inherent uncertainty, $\mathcal{SD}$ optimization clearly results in superior performance. This was further reflected in a smaller spread of the individual volume estimates (Figure~\ref{fig:boxplots}, Figure~\ref{fig:scale_specific_study_and_recalibration}-top, and the absolute differences in Table~\ref{tab:table_other_measures}). A possible explanation could be that the effects of the relative nature of the $\mathcal{SD}$ objective at a certain point trade off its induced volume bias. Furthermore, the larger spread of the individual data points after $\mathcal{CE}$ optimization could be linked with the experimental observation made by~\cite{Jungo2019} that although well-calibrated predictions could be observed at the dataset level, this may not be the case at the subject level.\\
Another interesting observation, based on Figure~\ref{fig:qualitative_inspection_ce_dice}-bottom, is that over-segmentation was observed rather than under-segmentation. In a way this confirmed the hypothesis that in realistic scenarios there are multiple regions with inherent uncertainty that are acting independently, which according to the simulations would result in easier over-estimation. We therefore believe that the intuitive example on creating independent regions on both sides of a tumor, due to observer variability at the borders of the tumor, was indeed valid. Nevertheless, it remains unclear how the reduction of model complexity relates to this, clearly re-enforcing this property.\\
It is not trivial to what extent the aleatoric or epistemic, respectively non-reducible and reducible, uncertainties play a role here. By reducing the complexity of the models one could argue that the simplest model increased the aleatoric uncertainty due to making the task descriptions more ambiguous, by only allowing intensity based classification. Instead, the epistemic uncertainty would increase due to an increased model uncertainty. Also, the extent of this is different between the different tasks. For IS17 and IS18 for example, on the one hand the availability of perfusion parameter maps requires the models to be less complex, while the limited number of cases does increase the approximation uncertainty significantly. We find it interesting that in machine learning indeed these two sources of uncertainty are usually not distinguished~\cite{Hullermeier2021}. State-of-the-art models often seem to be of sufficient complexity to reduce the overall inherent uncertainty to an extent to allow soft Dice optimization even for volume estimation.
\paragraph{Limitations and future work}
In this study, the analysis was limited to full-image processing. We therefore had to reduce the isotropic pixel- or voxel-sizes and limited the complexity of U-Net L compared to state-of-the-art. Although we obtained satisfactory performance, it must be noted that patch-wise training may result in substantial improvements. However, this would inevitably lead to inconsistencies between the theoretical analysis and the experimental validation due to a divergence between the optimization objective at testing time compared to the one used at training time. In this respect, it may be interesting to investigate the linear combination of $\mathcal{CE}$ and $\mathcal{SD}$, which is often seen to work well for patch-wise training~\cite{Isensee2018}, but further obscures the de facto optimization objective used during training and due to Pareto optimality will only trade-off the optimization objective.\\
It was also this full-image strategy that required an encoder-decoder based architecture, for which U-Net was the obvious choice, for all experiments. Nonetheless, we expect our findings to generalize across different CNN architectures (e.g. DeepMedic~\cite{Kamnitsas2017}, DeepLab~\cite{Chen2016}) since different architectures are used across top-competing methods (e.g.~\cite{Simpson2019}). They all seem to minimize the desired objective at testing time to a certain extent, thereby resulting in empirical risk minimization. A similar reasoning can be made for multi-class experiments. Instead, we opted to devote computation time as to perform an experimental validation across different complexities (i.e. LR, U-Net S, U-Net M and U-Net L) to virtually add inherent uncertainty and to obtain an idea of the necessity on the assumption of having perfect segmentation algorithms (as in Section~\ref{sec:theoretical_analysis}).\\
Another interesting line of research would be to analyze if objectives other than volume can also benefit from this kind of analysis. For example, in the segmentation of hippocampi, shape descriptors are often of particular interest as biomarkers since they seem to correlate well with particular types of diseases~\cite{Lindberg2012}. And from the side of training objective, other loss functions may be analyzed. However, $\mathcal{CE}$ and $\mathcal{SD}$ deserve special attention since they are most commonly used for segmentation tasks in the field of medical image analysis~\cite{Bertels2019b,Eelbode2020}.\\
Finally, the initial idea of comparing the pair-wise estimates and the concurrent hyper-parameter tuning of different methods led to using cross-validation across the entire datasets. When re-calibration was introduced this restricted the re-calibration parameters to be calculated from the training data, rather than from the validation data itself. It is expected that the volume-specific bias of the validation set will match the volume-specific bias of an independent test set more closely, due to possible overfitting on the training set. We therefore propose further investigation of re-calibration on validation data in situations where one has access to an independent test set and expect this type of re-calibration to deliver superior results.
\section{Conclusion}
In this work, different strategies to obtain a structure's volume from a segmentation map in the context of inherent uncertainty were presented. A theoretical analysis shed light on the appearance of segmentation maps after the application of different loss functions, namely cross-entropy and soft Dice. For cross-entropy, correct volume estimations are obtained via the direct summation of the soft segmentation maps, while soft Dice optimization results in binary-like segmentation maps with a potential volume bias proportional to the apparent inherent uncertainty. These findings were confirmed experimentally. However, with respect to volume estimation, a smaller spread on the individual data points and a superior performance on relative measures was observed in favor of soft Dice optimization. The higher complexity of state-of-the-art models reduces the inherent uncertainty to an extent allowing soft Dice optimization even when volume estimation is concerned. It was further shown that both optimizations still benefit from an ad-hoc volume analysis, with a simple re-calibration procedure based on the training set, thereby reducing the remaining volume bias and the volume-specific errors significantly.
\section*{Acknowledgments}
J.B. is part of NEXIS (\url{www.nexis-project.eu}), a project that has received funding from the European Union's Horizon 2020 Research and Innovations Programme (Grant Agreement \#780026). D.R. is supported by an innovation mandate of Flanders Innovation and Entrepreneurship (VLAIO).
\bibliographystyle{plain}
\bibliography{PhD_modified}
\end{document}

%% file: figures/inherent_uncertain_regions_with_volumes.tex
\begin{figure}[t]
\newcommand\myscale{0.4}
\setlength\tabcolsep{0pt}
\centering
\begin{tabular}{ccc}
    $K=1$ & $K=4$ & $K=16$\\
    \includegraphics[scale=\scale,scale=\myscale]{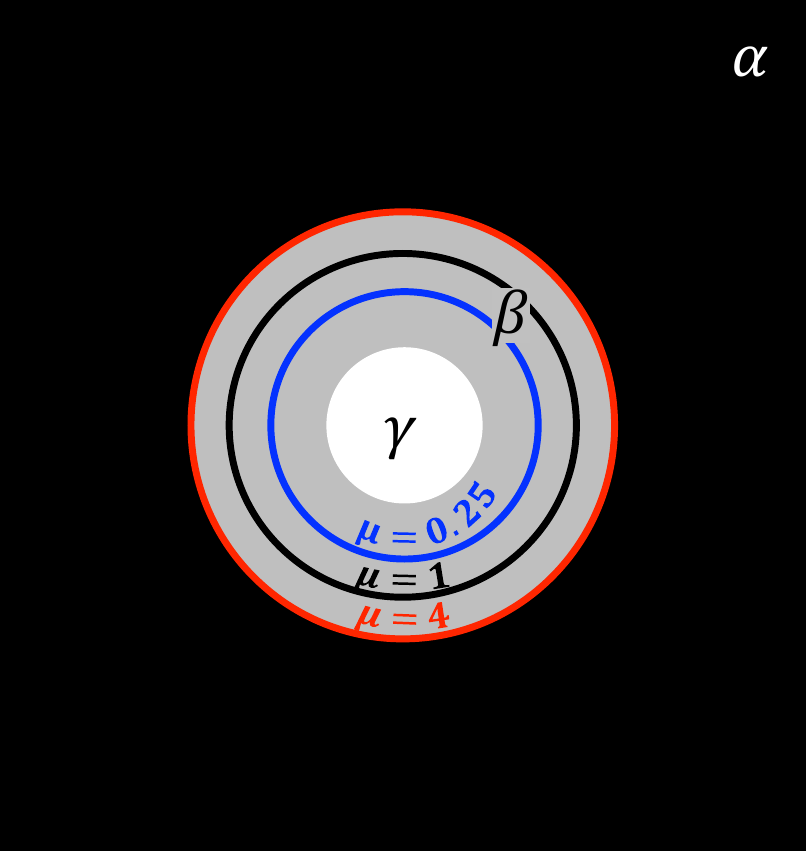}
    &\includegraphics[scale=\scale,scale=\myscale]{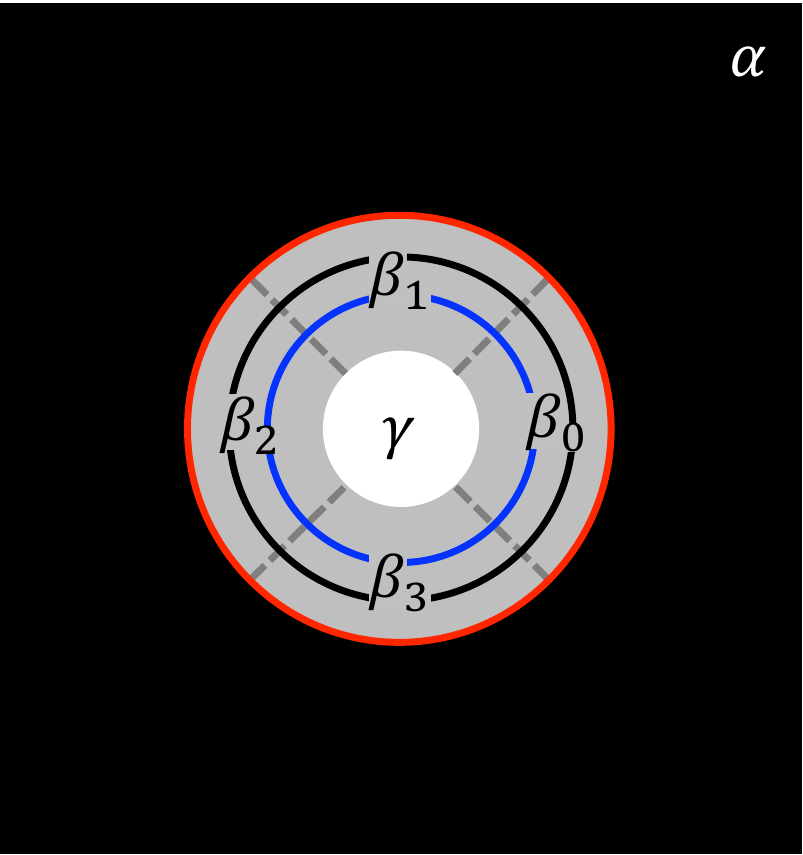}
    &\includegraphics[scale=\scale,scale=\myscale]{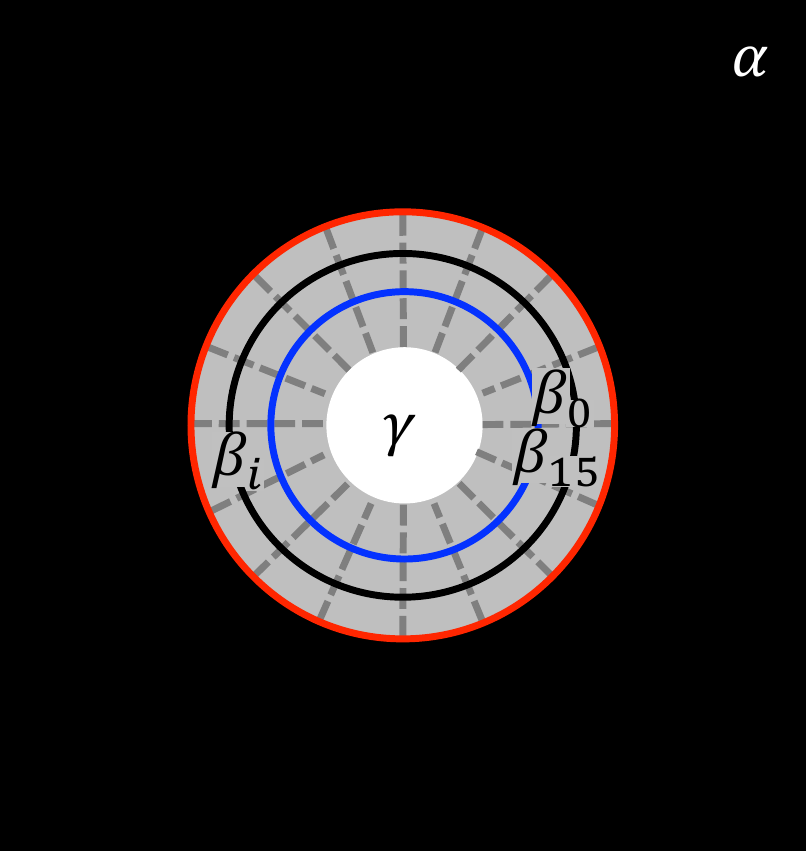}
\end{tabular}
\caption[Illustration of inherent uncertainty in a segmentation map.]{An illustration of inherent uncertainty in a segmentation map and the specific scenarios used for numerical simulation for $\mathcal{SD}$. The black area $\alpha$ corresponds to background with a foreground probability $p_\alpha$ of 0. The gray area $\beta$ (left figure) or areas $\beta_i$ (middle and right figure) are inherently uncertain and conditionally independent regions are separated by a dashed line. These regions belong to foreground with probabilities $p_\beta$ or $p_{\beta_i}$, respectively. The white area $\gamma$ corresponds to the structure with a foreground probability $p_\gamma$ of 1. Colored circles illustrate the extent (size) of the total inherently uncertain area.}
\label{fig:inherent_uncertain_regions_with_volumes}
\end{figure}

%% file: figures/all_regions_plot.tex
\begin{figure}[t]
\newcommand\myscale{0.17}
\setlength\tabcolsep{0pt}
\centering
\begin{tabular}{cccc}
    &\enspace $K=1$ & \enspace $K=4$ & \enspace $K=16$\\
    \rotatebox[origin=c]{90}{$\mathbb{E}[\mathcal{SD}]$}
    &\raisebox{-0.5\height}{\includegraphics[scale=\scale,scale=\myscale]{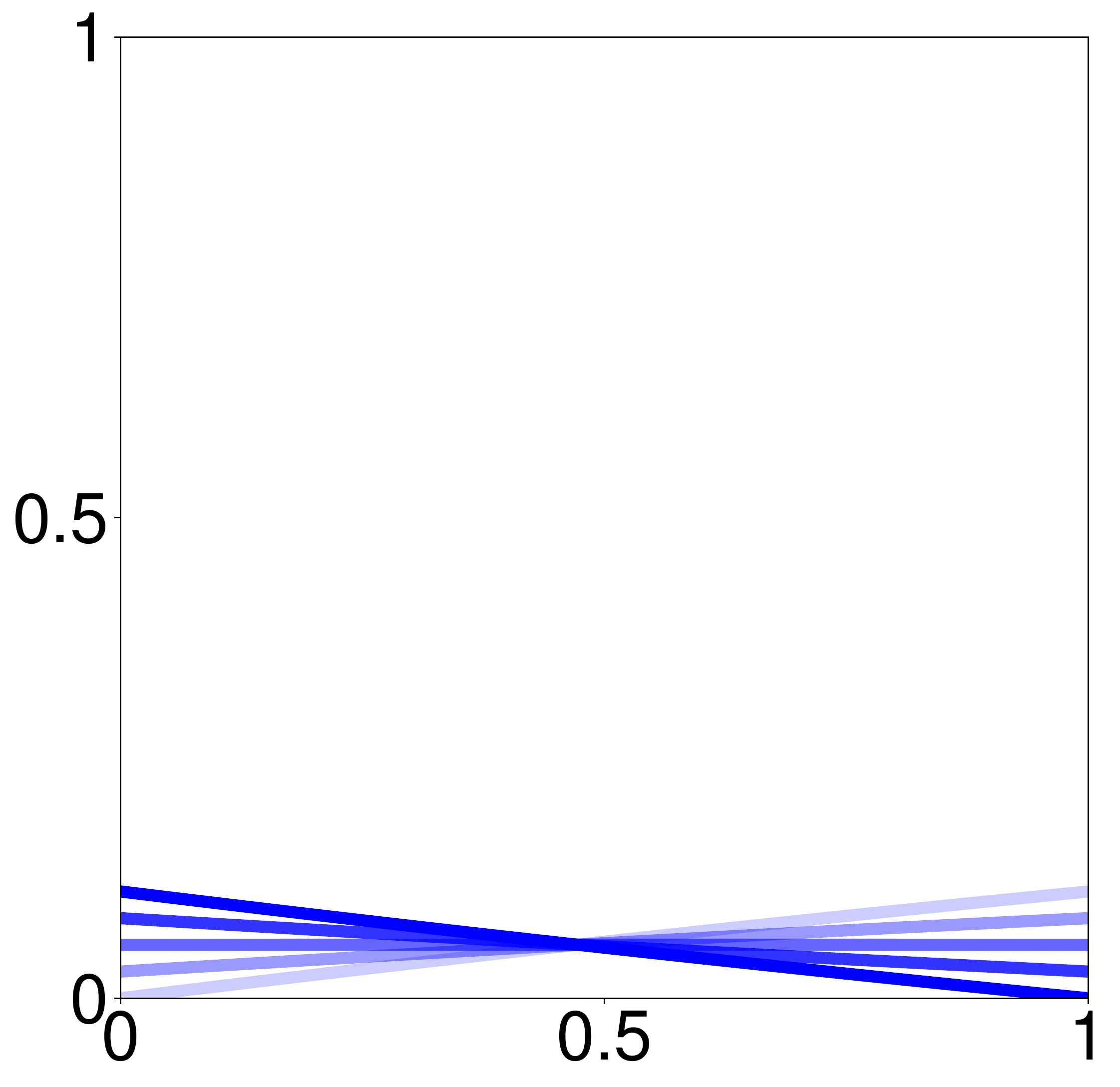}}
    &\raisebox{-0.5\height}{\includegraphics[scale=\scale,scale=\myscale]{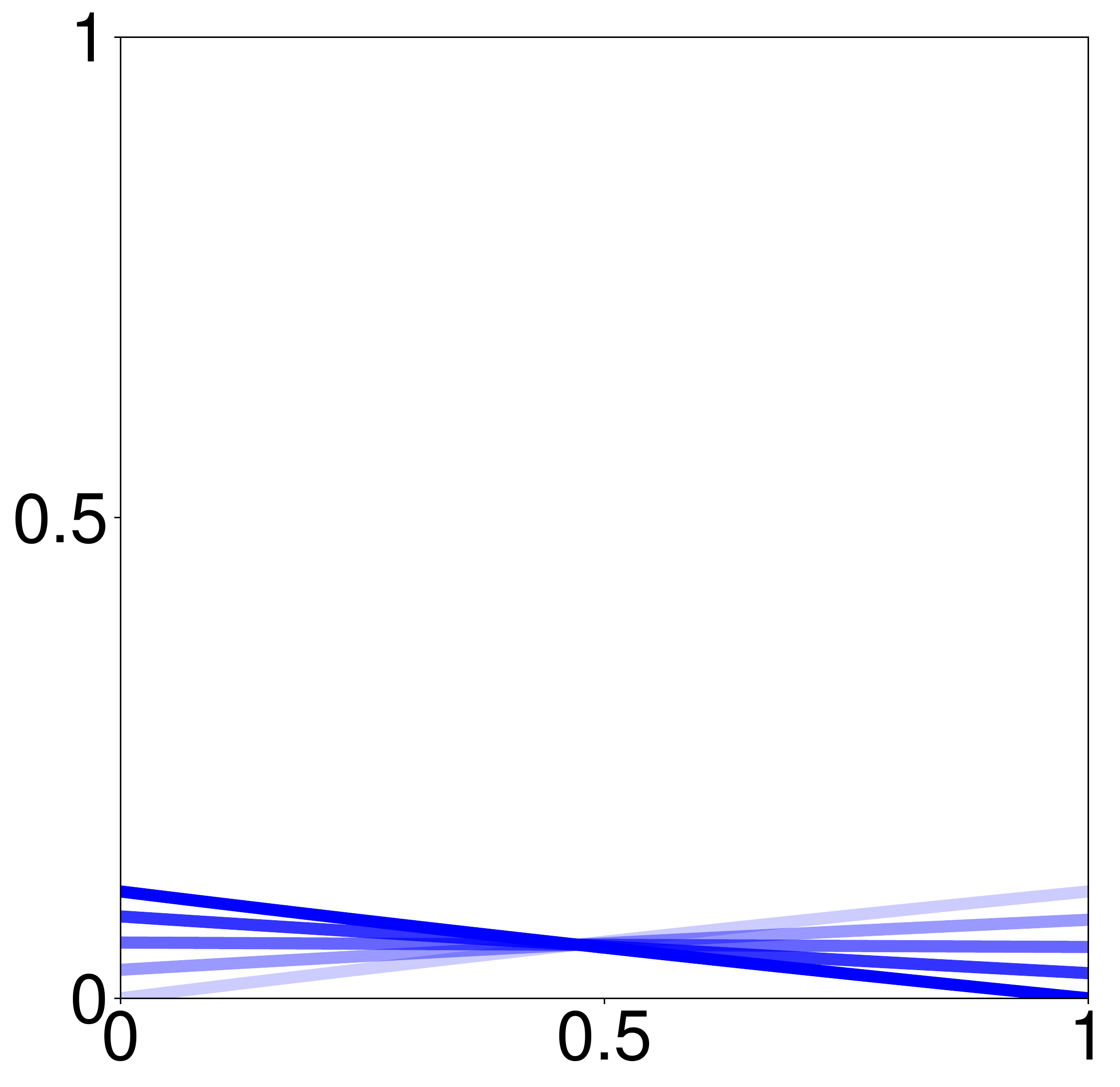}}
    &\raisebox{-0.5\height}{\includegraphics[scale=\scale,scale=\myscale]{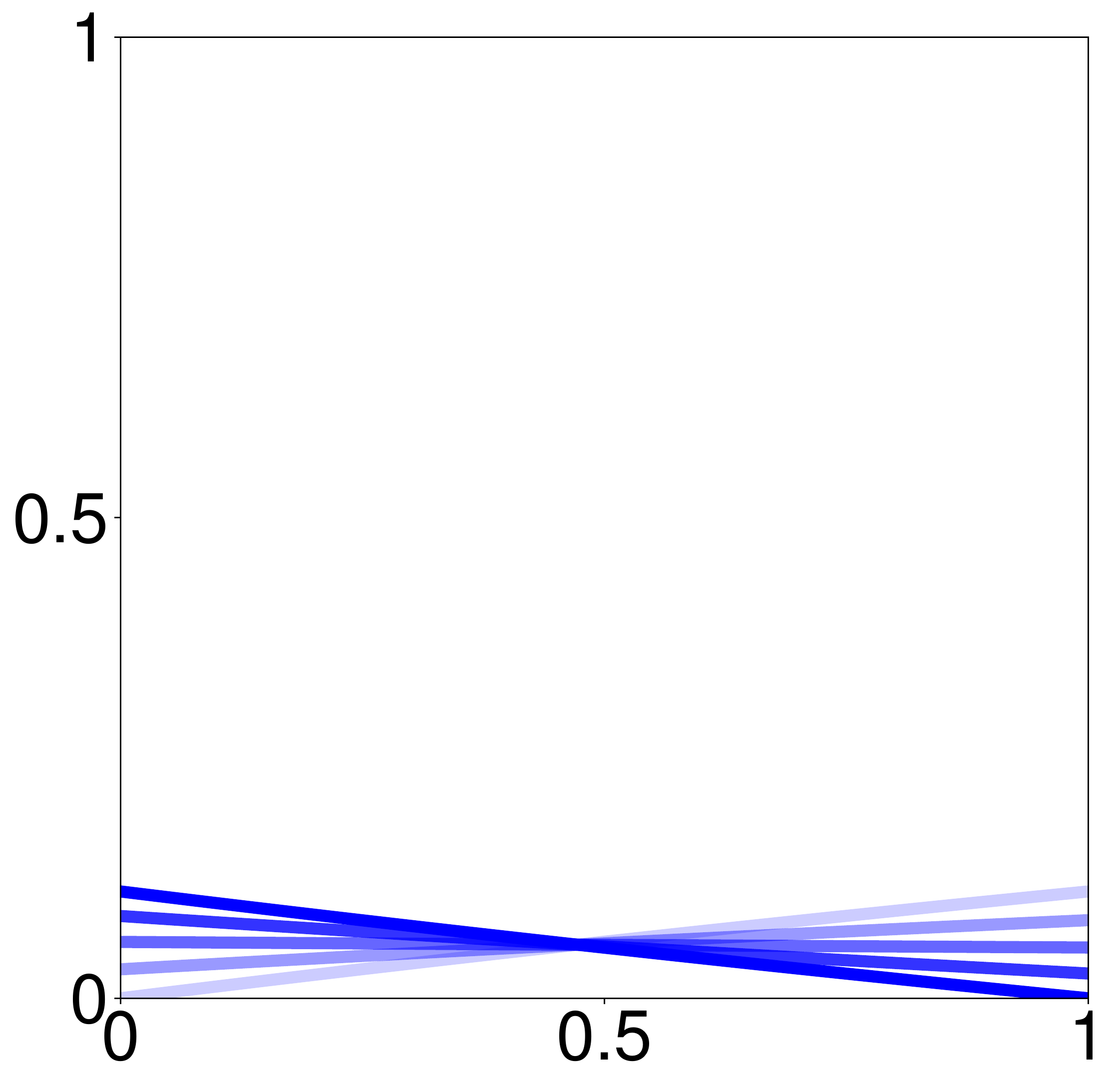}}\\
    \rotatebox[origin=c]{90}{$\mathbb{E}[\mathcal{SD}]$}
    &\raisebox{-0.5\height}{\includegraphics[scale=\scale,scale=\myscale]{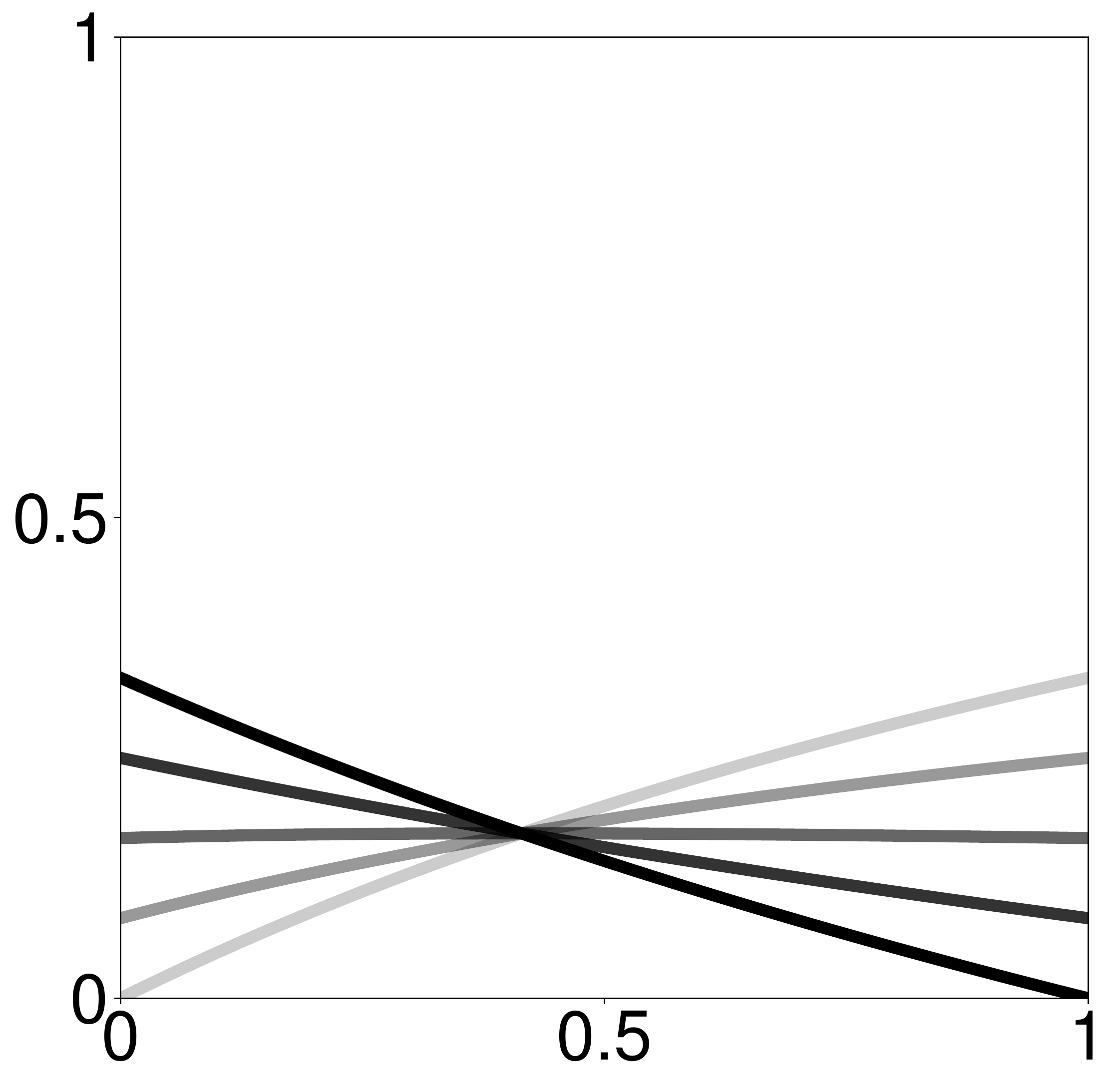}}
    &\raisebox{-0.5\height}{\includegraphics[scale=\scale,scale=\myscale]{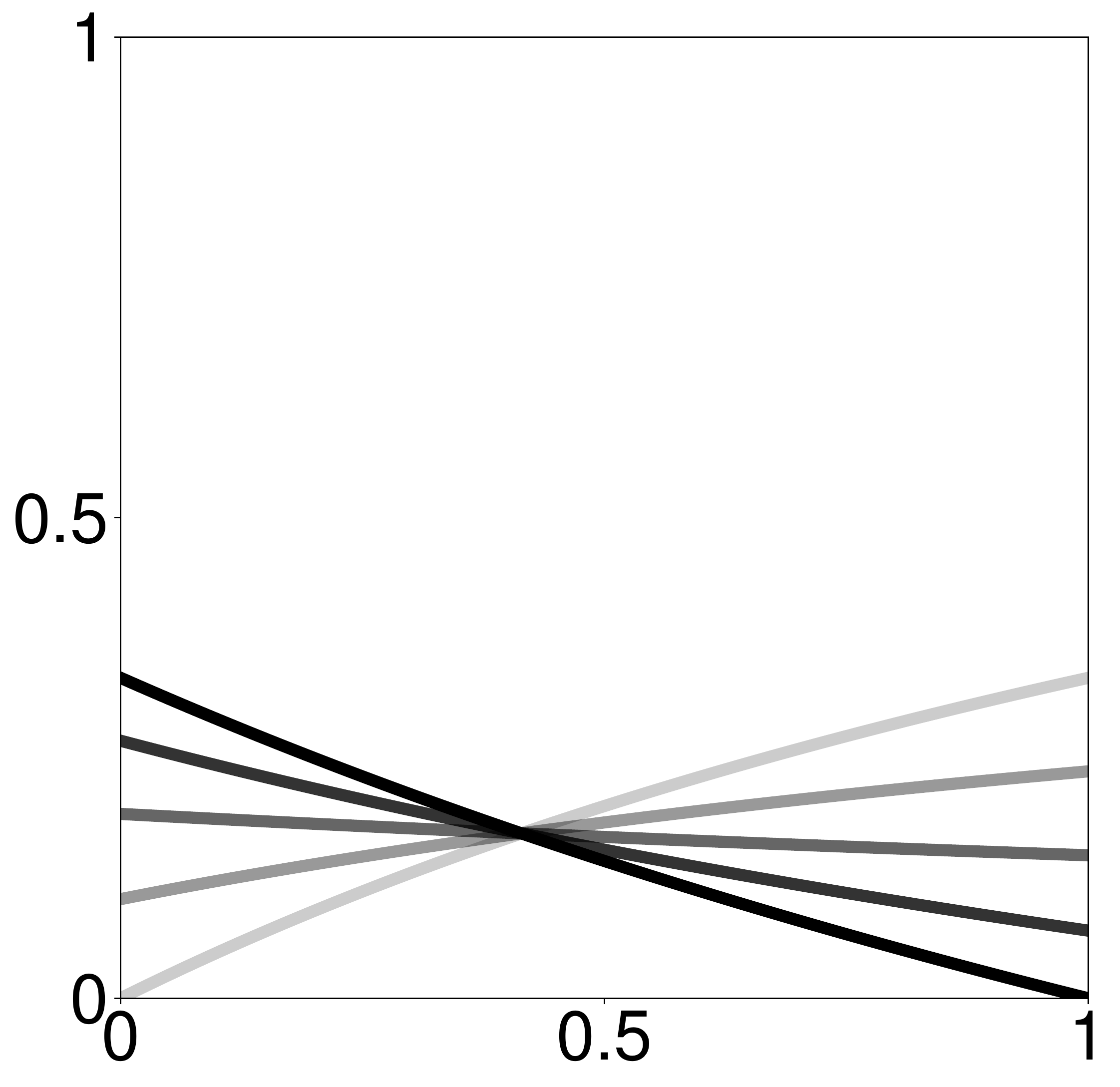}}
    &\raisebox{-0.5\height}{\includegraphics[scale=\scale,scale=\myscale]{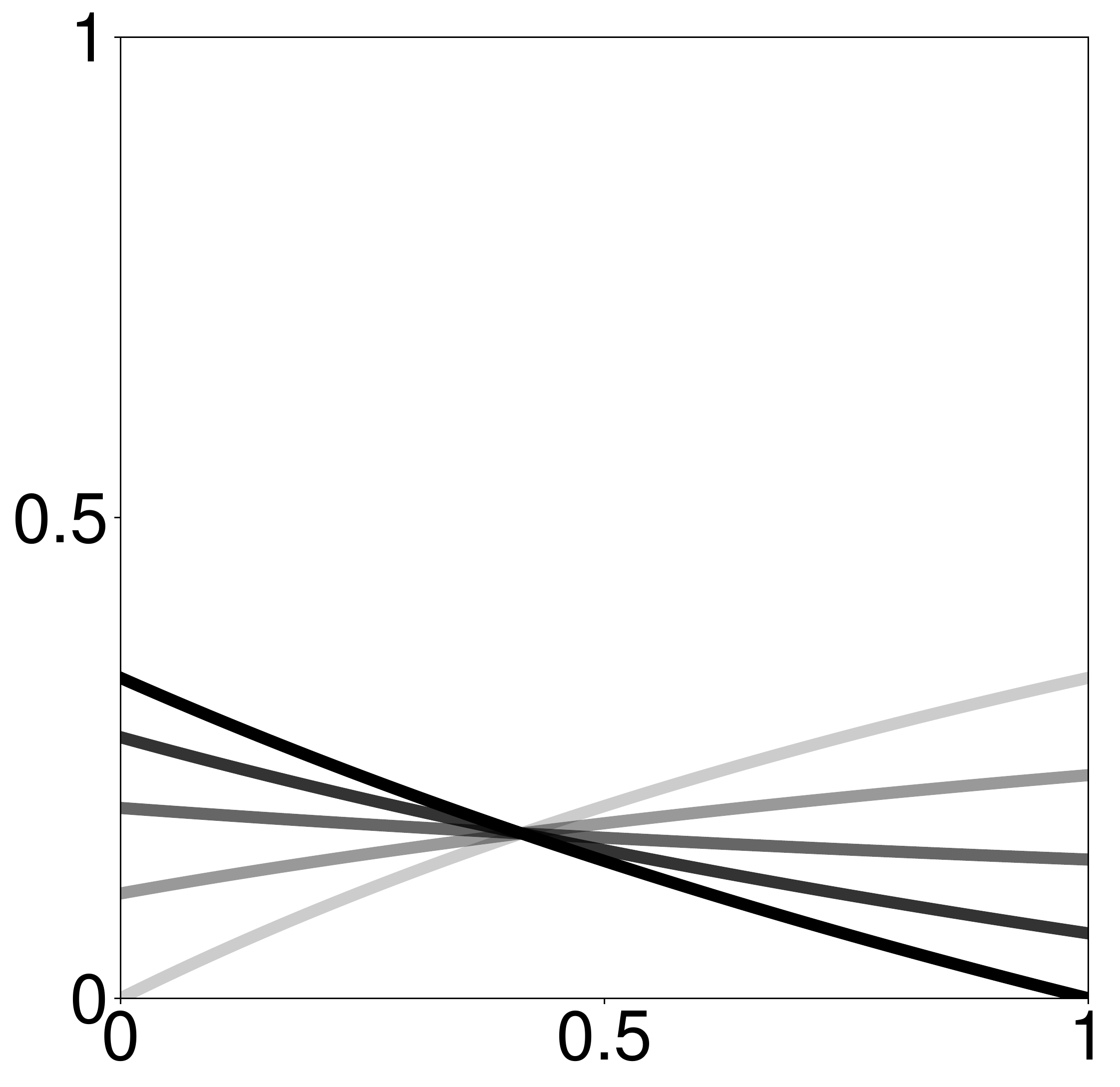}}\\
    \rotatebox[origin=c]{90}{$\mathbb{E}[\mathcal{SD}]$}
    &\raisebox{-0.5\height}{\includegraphics[scale=\scale,scale=\myscale]{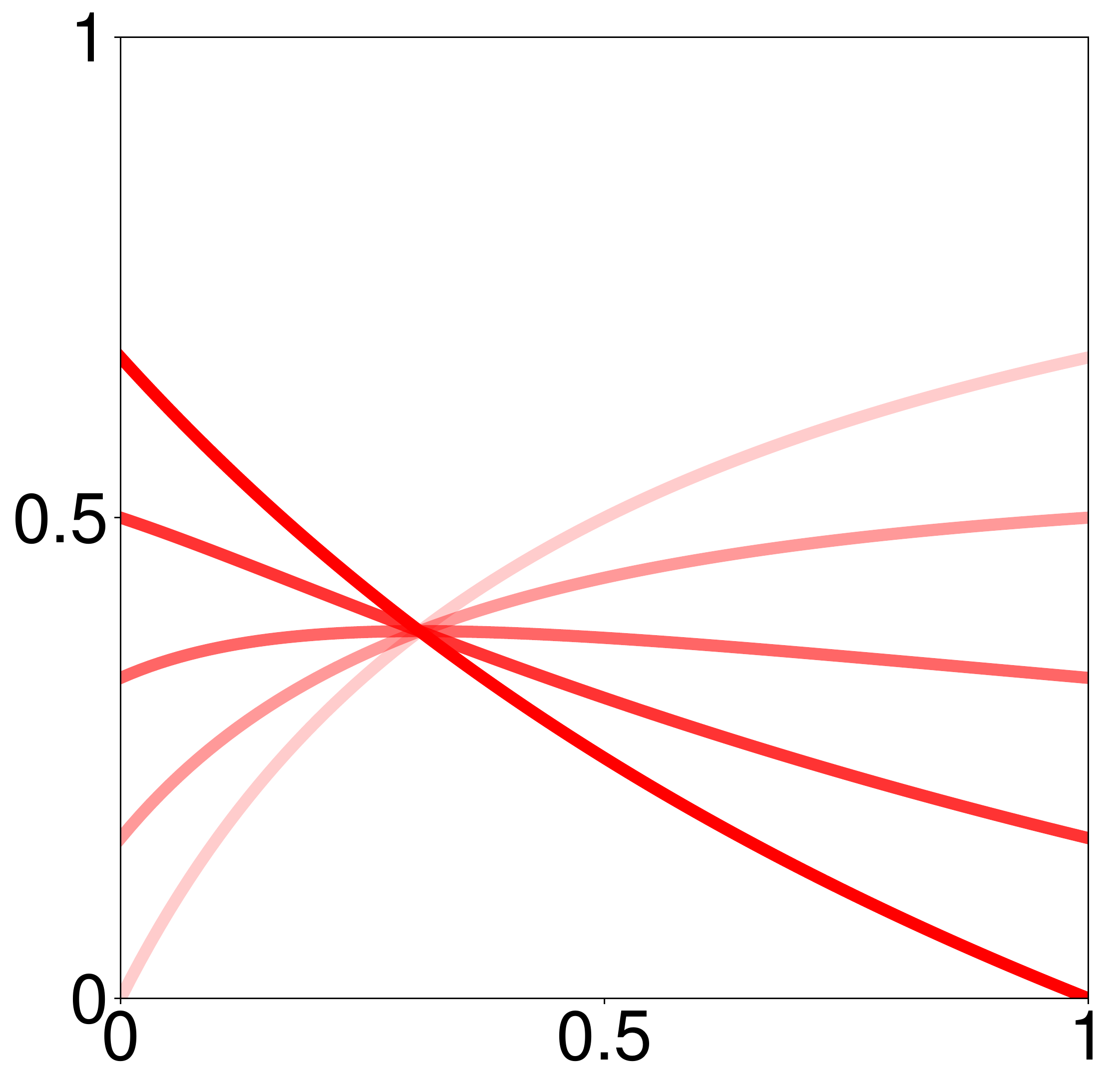}}
    &\raisebox{-0.5\height}{\includegraphics[scale=\scale,scale=\myscale]{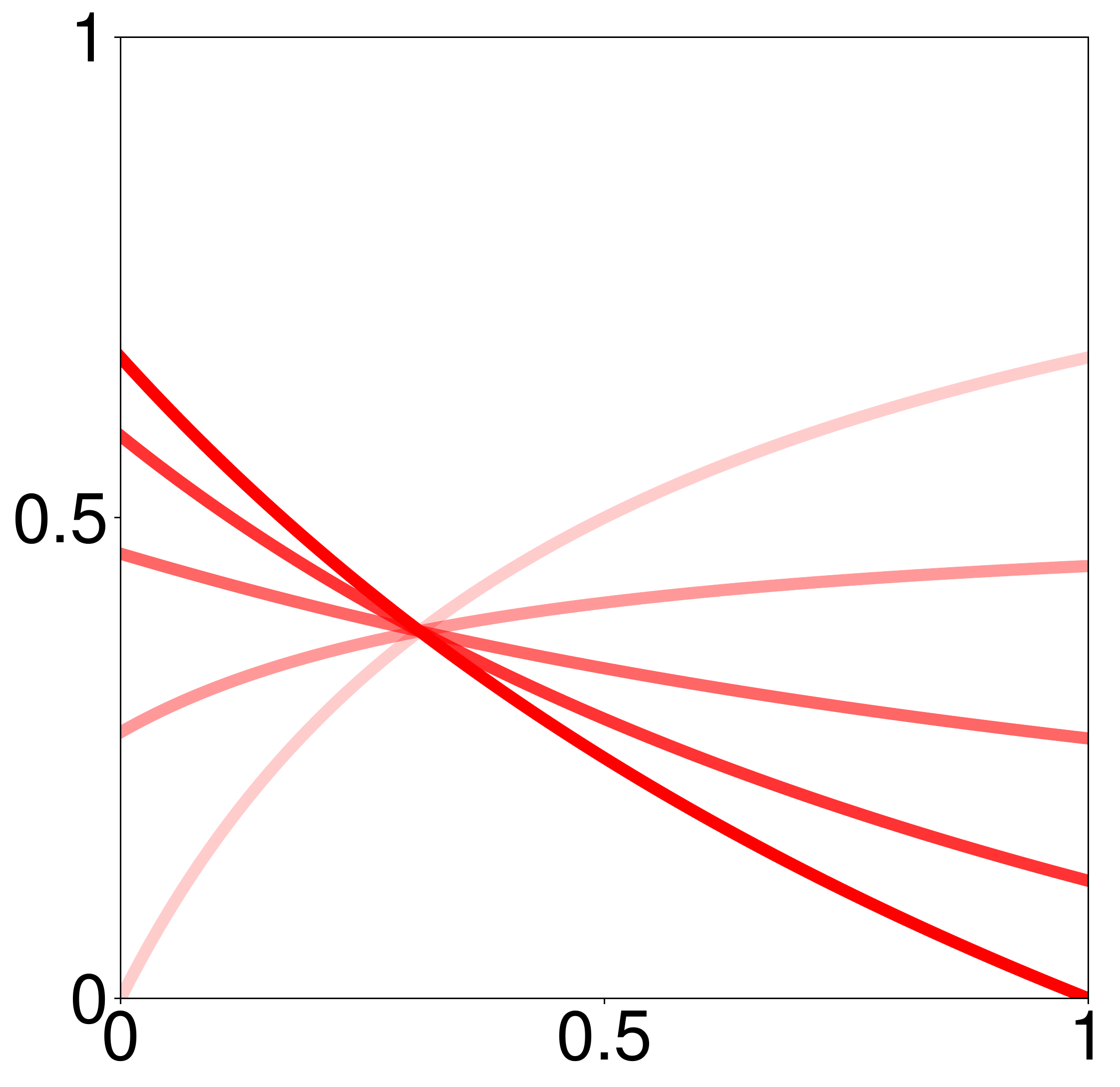}}
    &\raisebox{-0.5\height}{\includegraphics[scale=\scale,scale=\myscale]{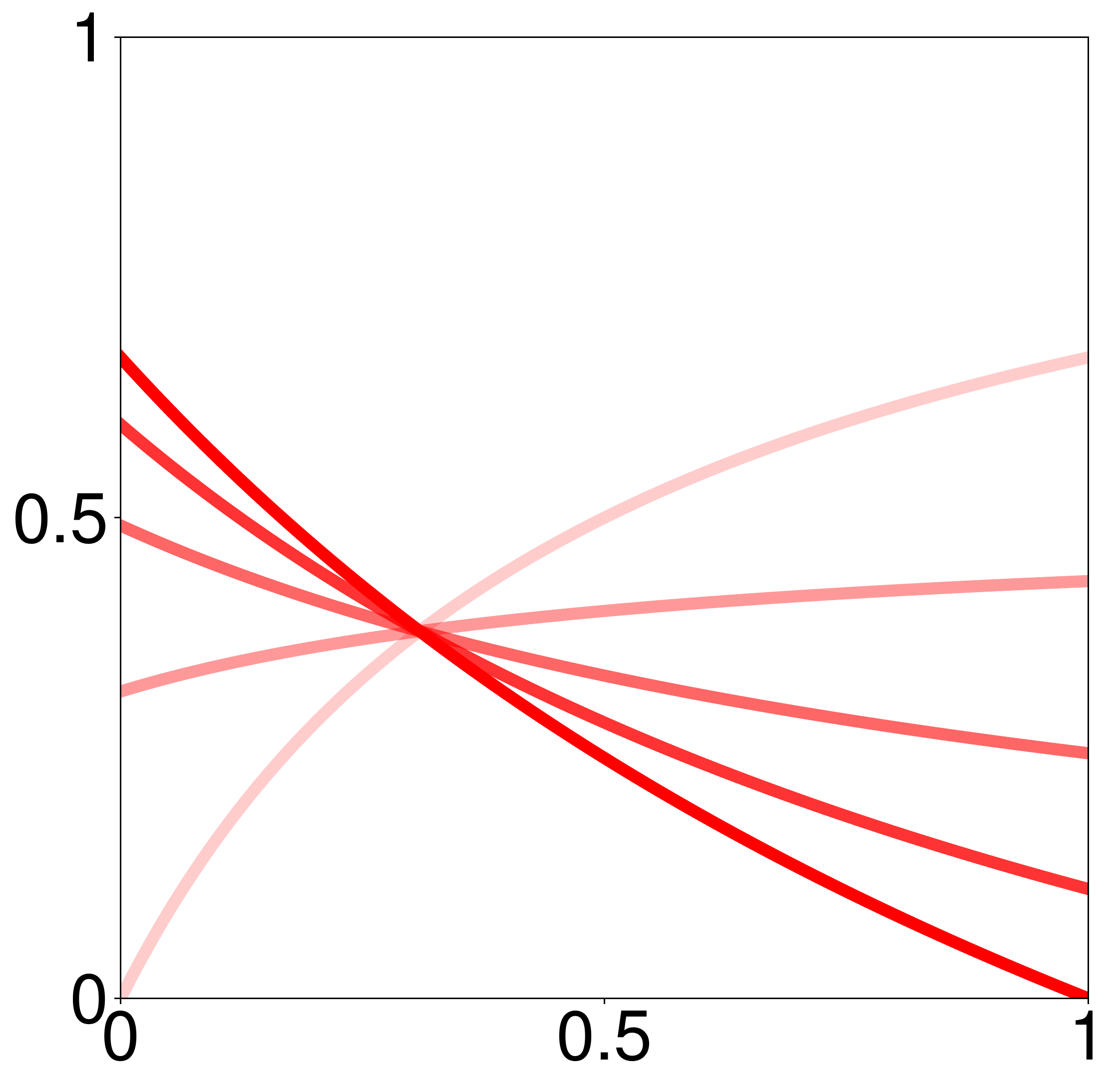}}\\
    &\enspace $\tilde{p}_\beta$ & \enspace $\tilde{p}_\beta$ & \enspace $\tilde{p}_\beta$
\end{tabular}
\caption[The expected $\mathcal{SD}$ loss value as a function of the predicted foreground probability for different true foreground probabilities.]{The expected $\mathcal{SD}$ loss value as a function of the predicted foreground probability $\tilde{p}_\beta$ for different true foreground probabilities $p_\beta$ (a higher opacity points to a higher true foreground probability $p_\beta=\{0, 0.25, 0.5, 0.75, 1\}$). Different rows/colors represent different total volumes of the uncertain area: $\mu=0.25$ (blue), $\mu=1$ (black), $\mu=4$ (red). Numerical results are given for K=\{1, 4, 16\} independent regions, respectively left to right column.}
\label{fig:all_regions_plot}
\end{figure}

%% file: figures/p_bias_plot.tex
\begin{figure}[t]
\newcommand\myscale{0.17}
\setlength\tabcolsep{0pt}
\centering
\begin{tabular}{cccc}
&\quad $K=1$ & \quad $K=4$ & \quad $K=16$\\
\rotatebox[origin=c]{90}{$\tilde{p}_\beta-p_\beta$}
&\raisebox{-0.5\height}{\includegraphics[scale=\scale,scale=\myscale]{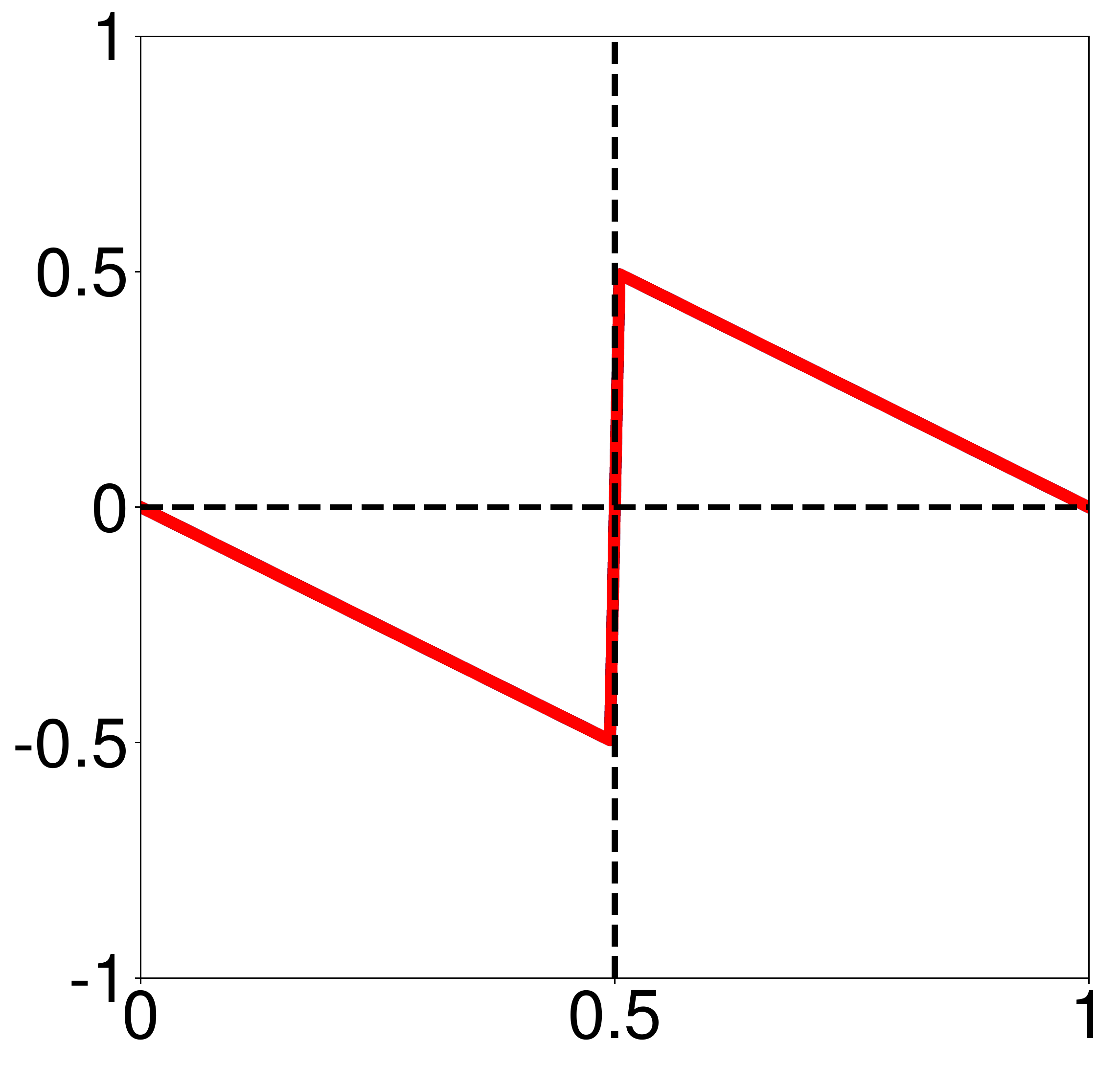}}
&\raisebox{-0.5\height}{\includegraphics[scale=\scale,scale=\myscale]{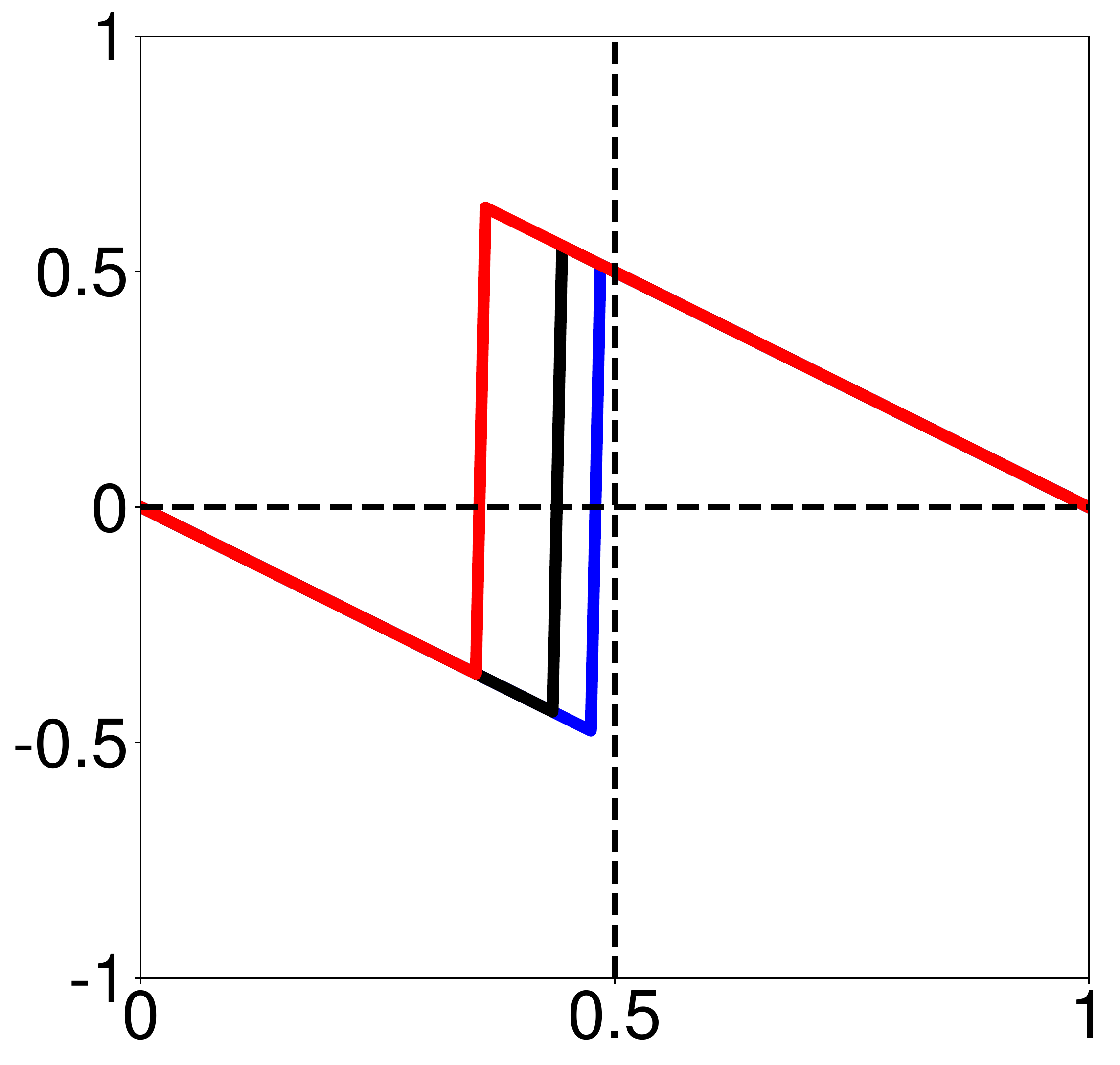}}
&\raisebox{-0.5\height}{\includegraphics[scale=\scale,scale=\myscale]{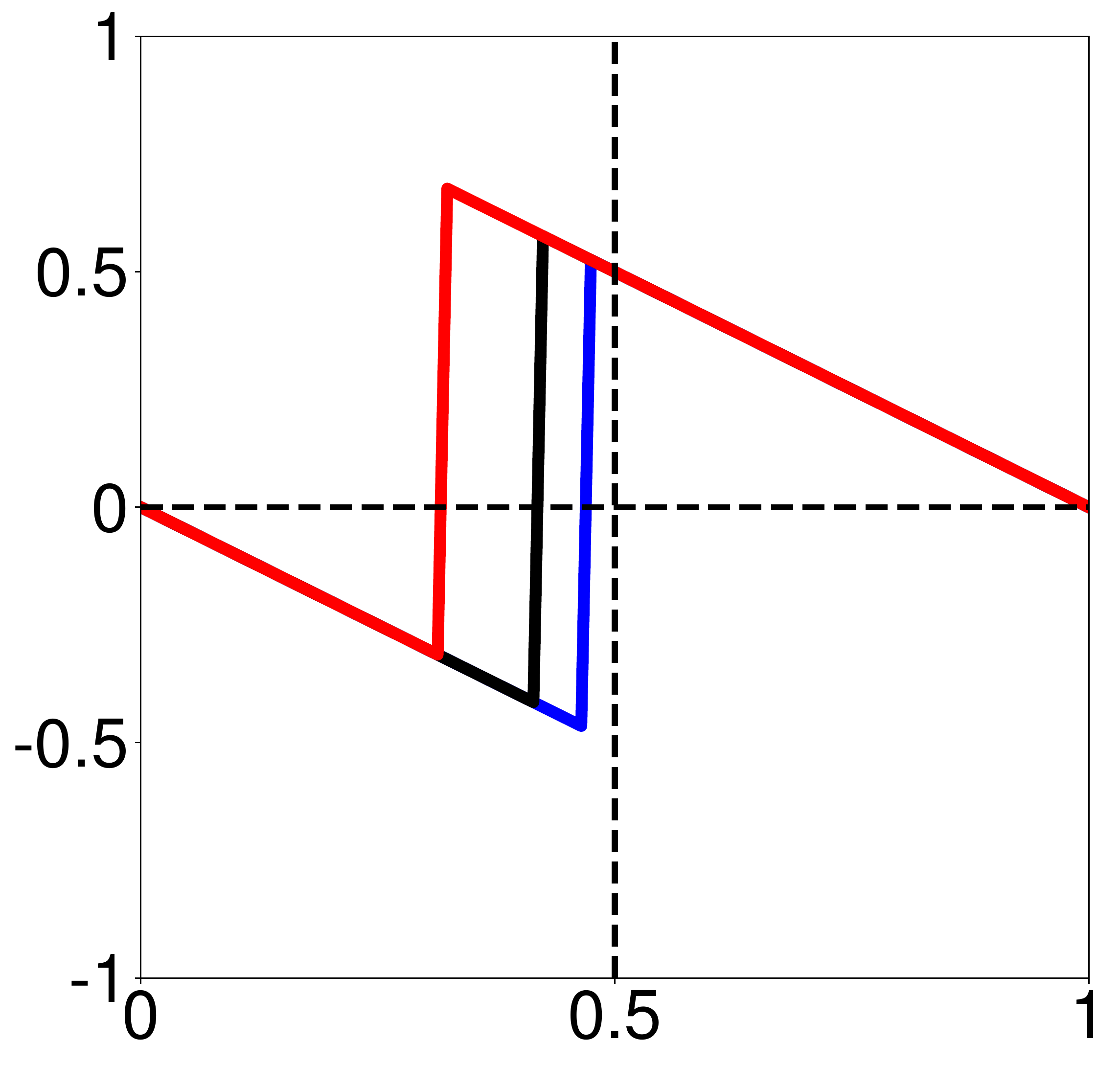}}\\
&\quad $p_\beta$ & \quad $p_\beta$ & \quad $p_\beta$
\end{tabular}
\caption[Error on the predicted foreground probability as a function of the true foreground probability after $\mathcal{SD}$ optimization.]{Error on the predicted foreground probability as a function of the true foreground probability $p_\beta$ after $\mathcal{SD}$ optimization. Different colors represent different total volumes of the uncertain area: $\mu=0.25$ (blue), $\mu=1$ (black), $\mu=4$ (red). Numerical results are given for K=\{1, 4, 16\} independent regions, respectively left to right column.}
\label{fig:p_bias_plot}
\end{figure}

%% file: figures/networks.tex
\begin{figure}[t]
\newcommand\myscale{0.65}
\setlength\tabcolsep{0pt}
\centering
\includegraphics[scale=\scale,scale=\myscale]{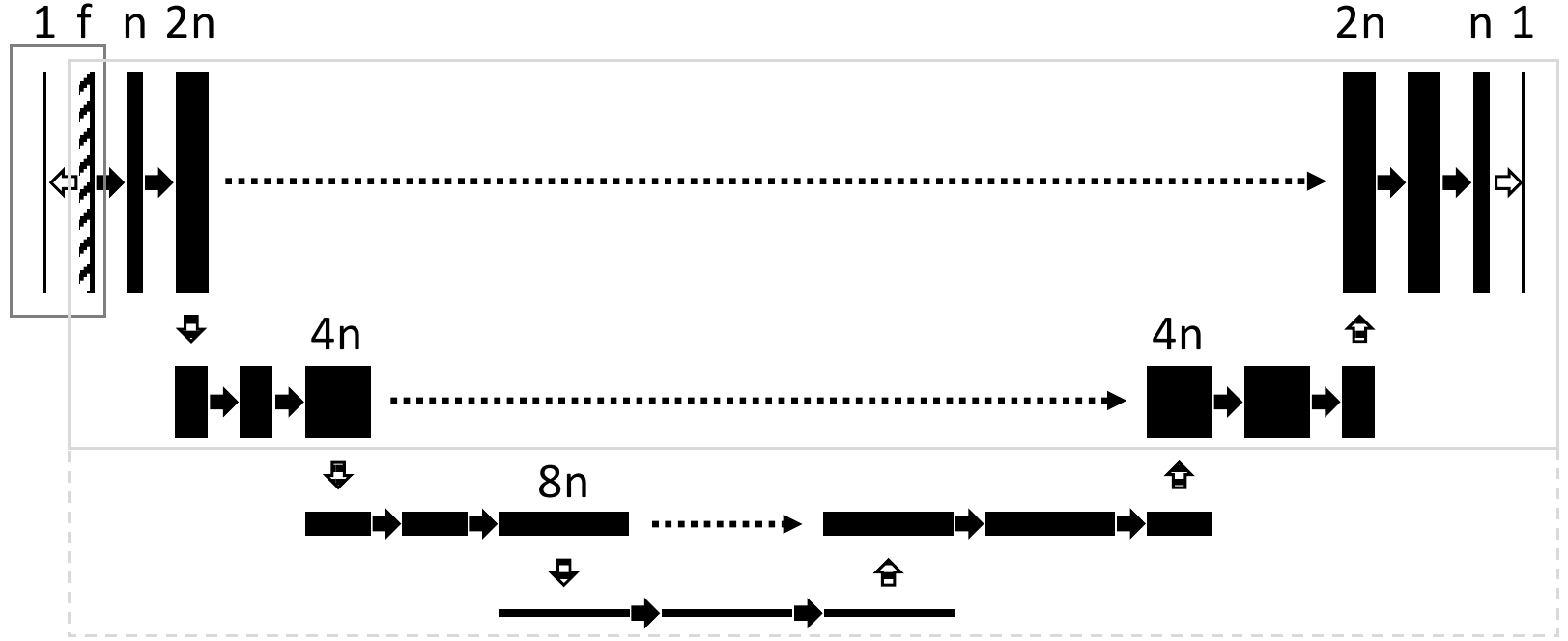}
\caption[Network architectures.]{Network architectures. LR model in dark gray rectangle. U-Net S model in light gray rectangle. For U-Net M and L models this extends up until the dashed line. Legend: f - \# input features (dataset dependent)); n - \# feature maps; unfilled horizontal arrow - 1x1(x1) convolution with sigmoid activation; filled horizontal arrow - 3x3(x3) convolution with leaky-ReLU activation; down arrow - average/max-pooling; up arrow - bi/tri-linear upsampling; dashed horizontal arrow: alignment cropping and concatenation.}
\label{fig:networks}
\end{figure}

%% file: tables/table_volumetric_bias.tex
\begin{table}[t]
\setlength{\tabcolsep}{3pt}
\centering
\caption[The volume bias after $\mathcal{CE}$ or $\mathcal{SD}$ optimization for four models trained on four medical tasks.]{The volume bias $\mathbb{E}[\Delta\mathcal{V}]$ after $\mathcal{CE}$ or $\mathcal{SD}$ optimization for the four models trained on four medical tasks. Values closer to zero are better and bold or italic are values significantly greater or smaller than zero. MO17 and BR18 have relatively low inherent uncertainty, and IS17 and IS18 have relatively high inherent uncertainty. In terms of bias, after $\mathcal{CE}$ optimization the volumes of the fuzzy predictions $\mathcal{V}(\tilde{y})$ should be used. After $\mathcal{SD}$ optimization the effect is similar for $\mathcal{V}(\tilde{y})$ and for $\mathcal{V}(\tilde{l})$, the volumes after thresholding at 0.5.}
\label{tab:table_volumetric_bias}
\begin{tabular}{ll|rr|rr|rr|rr}
\toprule
& \multicolumn{1}{r|}{Model $\rightarrow$} & \multicolumn{2}{c|}{LR} & \multicolumn{2}{c|}{U-Net S} & \multicolumn{2}{c|}{U-Net M} & \multicolumn{2}{c}{U-Net L} \\
& \multicolumn{1}{r|}{Training loss $\rightarrow$} & \multicolumn{1}{c}{$\mathcal{CE}$} & \multicolumn{1}{c|}{$\mathcal{SD}$} & \multicolumn{1}{c}{$\mathcal{CE}$} & \multicolumn{1}{c|}{$\mathcal{SD}$} & \multicolumn{1}{c}{$\mathcal{CE}$} & \multicolumn{1}{c|}{$\mathcal{SD}$} & \multicolumn{1}{c}{$\mathcal{CE}$} & \multicolumn{1}{c}{$\mathcal{SD}$} \\
Dataset $\downarrow$ & Metric $\downarrow$ & & & & & & \\
\midrule
\multirow{2}{*}{MO17} & $\mathbb{E}[\Delta\mathcal{V}(\tilde{y},l)] = \mathbb{E}[\mathcal{V}(\tilde{y})-\mathcal{V}(l)]$ ($10^3$ pxls) & -0.07 & \textbf{302.29} & -0.28 & \textbf{87.09} & 0.09 & -0.19 & -0.14 & 0.00 \\
     & $\mathbb{E}[\Delta\mathcal{V}(\tilde{l},l)] = \mathbb{E}[\mathcal{V}(\tilde{l})-\mathcal{V}(l)]$ ($10^3$ pxls) & \textit{-33.064} & \textbf{303.39} & \textit{-33.06} & \textbf{86.61} & 0.15 & -0.05 & -0.01 & 0.14 \\
\cline{1-10}
\multirow{2}{*}{BR18} & $\mathbb{E}[\Delta\mathcal{V}(\tilde{y},l)] = \mathbb{E}[\mathcal{V}(\tilde{y})-\mathcal{V}(l)]$ (ml) & -2.84 & \textbf{276.43} & 3.94 & \textbf{19.93} & \textit{-6.78} & -1.91 & \textit{-3.21} & \textit{-3.95} \\
     & $\mathbb{E}[\Delta\mathcal{V}(\tilde{l},l)] = \mathbb{E}[\mathcal{V}(\tilde{l})-\mathcal{V}(l)]$ (ml) & \textit{-96.30} & \textbf{256.92} & \textit{-84.78} & \textbf{19.32} & \textit{-14.47} & -1.98 & \textit{-8.94} & \textit{-3.90} \\
\cline{1-10}
\multirow{2}{*}{IS17} & $\mathbb{E}[\Delta\mathcal{V}(\tilde{y},l)] = \mathbb{E}[\mathcal{V}(\tilde{y})-\mathcal{V}(l)]$ (ml) & 15.71 & \textbf{82.42} & -4.23 & \textbf{23.84} & -2.88 & \textbf{13.44} & 0.22 & 2.46 \\
     & $\mathbb{E}[\Delta\mathcal{V}(\tilde{l},l)] = \mathbb{E}[\mathcal{V}(\tilde{l})-\mathcal{V}(l)]$ (ml) & -19.02 & \textbf{79.00} & \textit{-33.97} & \textbf{23.85} & \textit{-23.01} & \textbf{13.71} & \textit{-13.23} & 2.51 \\
\cline{1-10}
\multirow{2}{*}{IS18} & $\mathbb{E}[\Delta\mathcal{V}(\tilde{y},l)] = \mathbb{E}[\mathcal{V}(\tilde{y})-\mathcal{V}(l)]$ (ml) & 0.77 & \textbf{34.03} & -0.37 & \textbf{12.44} & -0.88 & \textbf{5.44} & \textbf{3.57} & \textbf{6.46} \\
     & $\mathbb{E}[\Delta\mathcal{V}(\tilde{l},l)] = \mathbb{E}[\mathcal{V}(\tilde{l})-\mathcal{V}(l)]$ (ml) & \textit{-24.56} & \textbf{33.09} & \textit{-24.69} & \textbf{12.22} & \textit{-6.80} & \textbf{5.47} & \textit{-6.22} & \textbf{6.37} \\
\bottomrule
\end{tabular}
\end{table}

%% file: figures/boxplots.tex
\begin{figure}[ht]
\newcommand\myscale{0.17}
\setlength\tabcolsep{0pt}
\centering
\begin{tabular}{ccccc}
&\quad MO17 & \quad BR18 & \quad IS17 & \quad IS18\\
\rotatebox[origin=c]{90}{$\Delta\mathcal{V}$ (ml or $10^3$ pixels)}
&\raisebox{-0.5\height}{\includegraphics[scale=\scale,scale=\myscale]{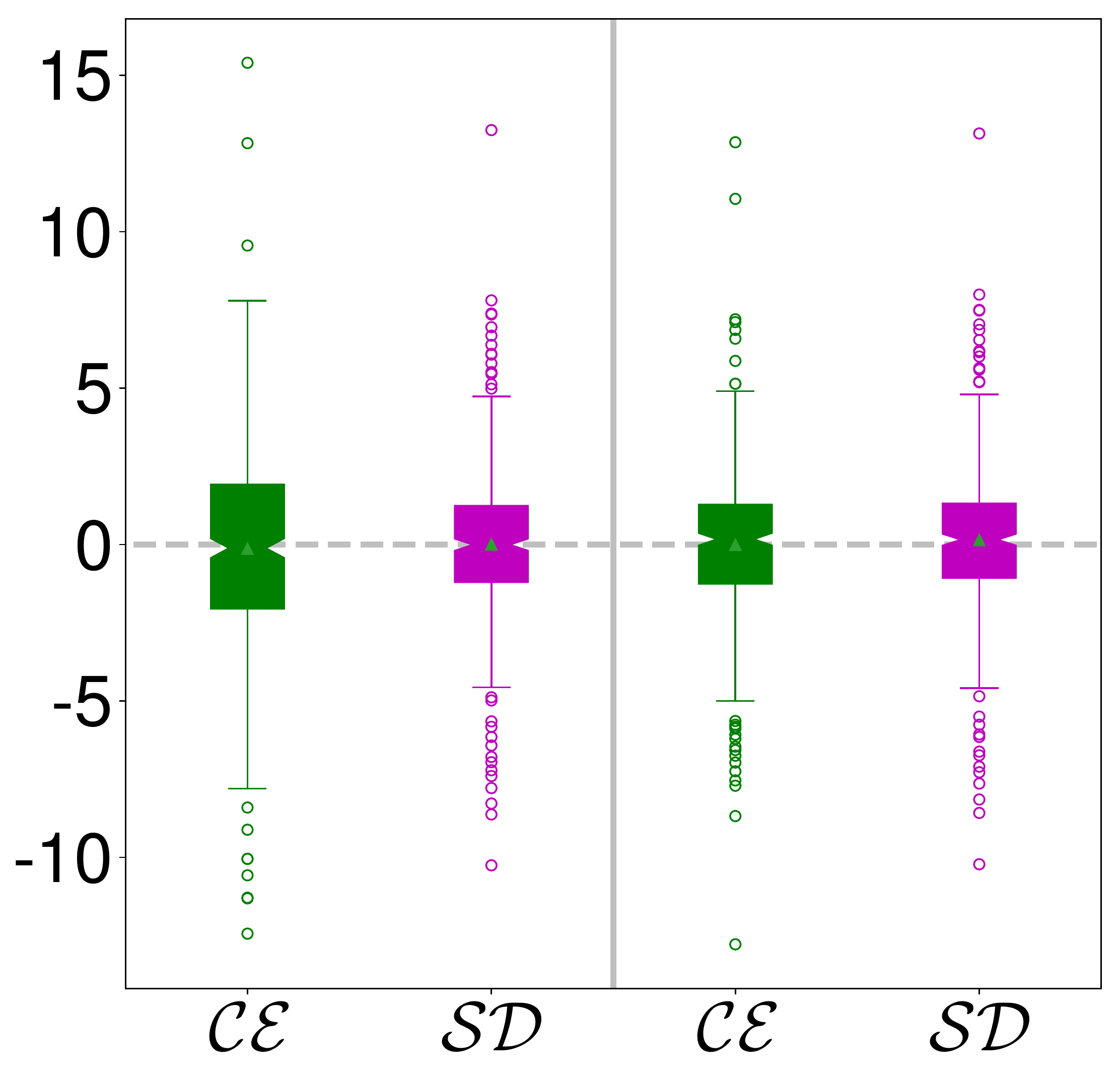}}
&\raisebox{-0.5\height}{\includegraphics[scale=\scale,scale=\myscale]{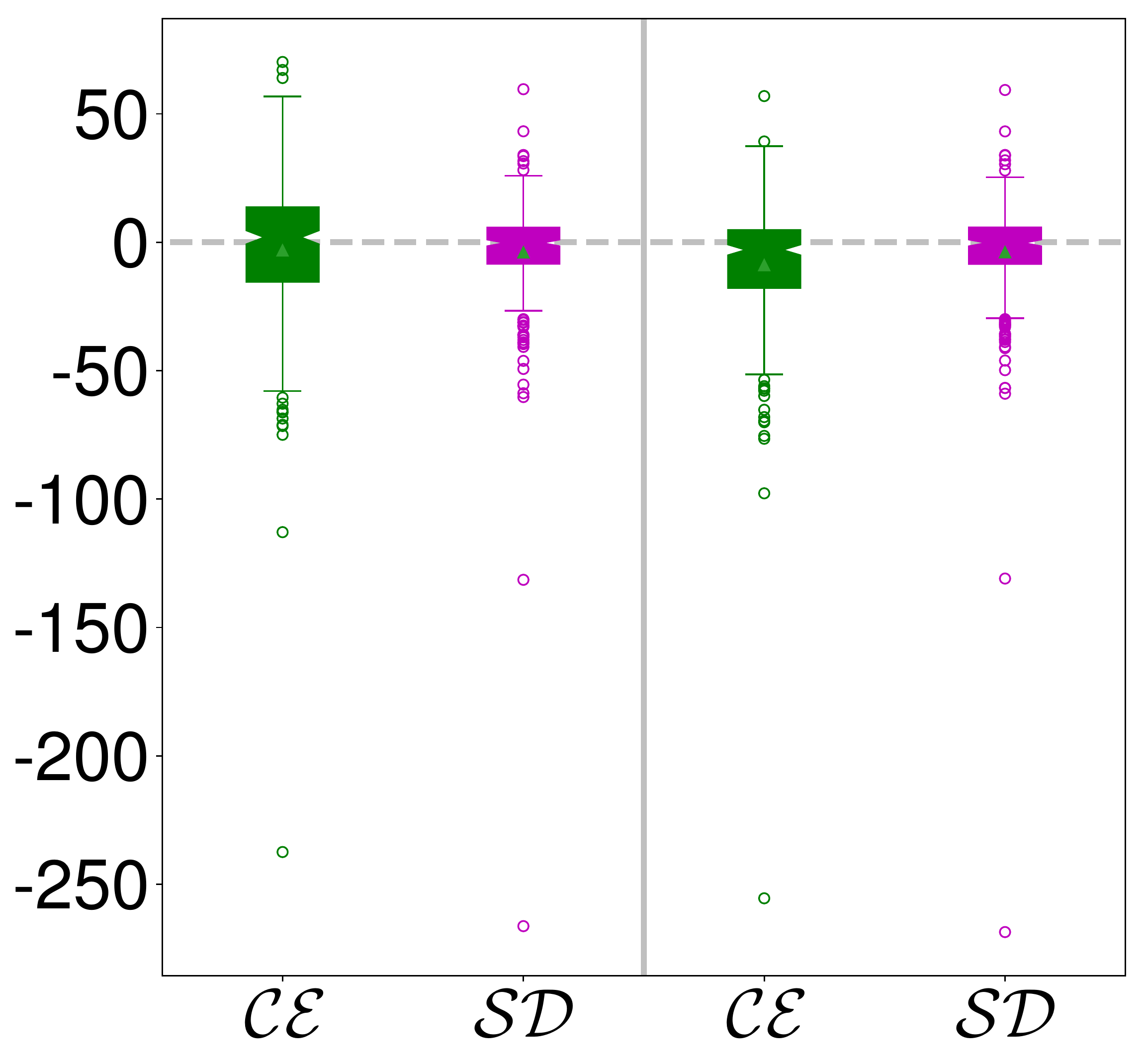}}
&\raisebox{-0.5\height}{\includegraphics[scale=\scale,scale=\myscale]{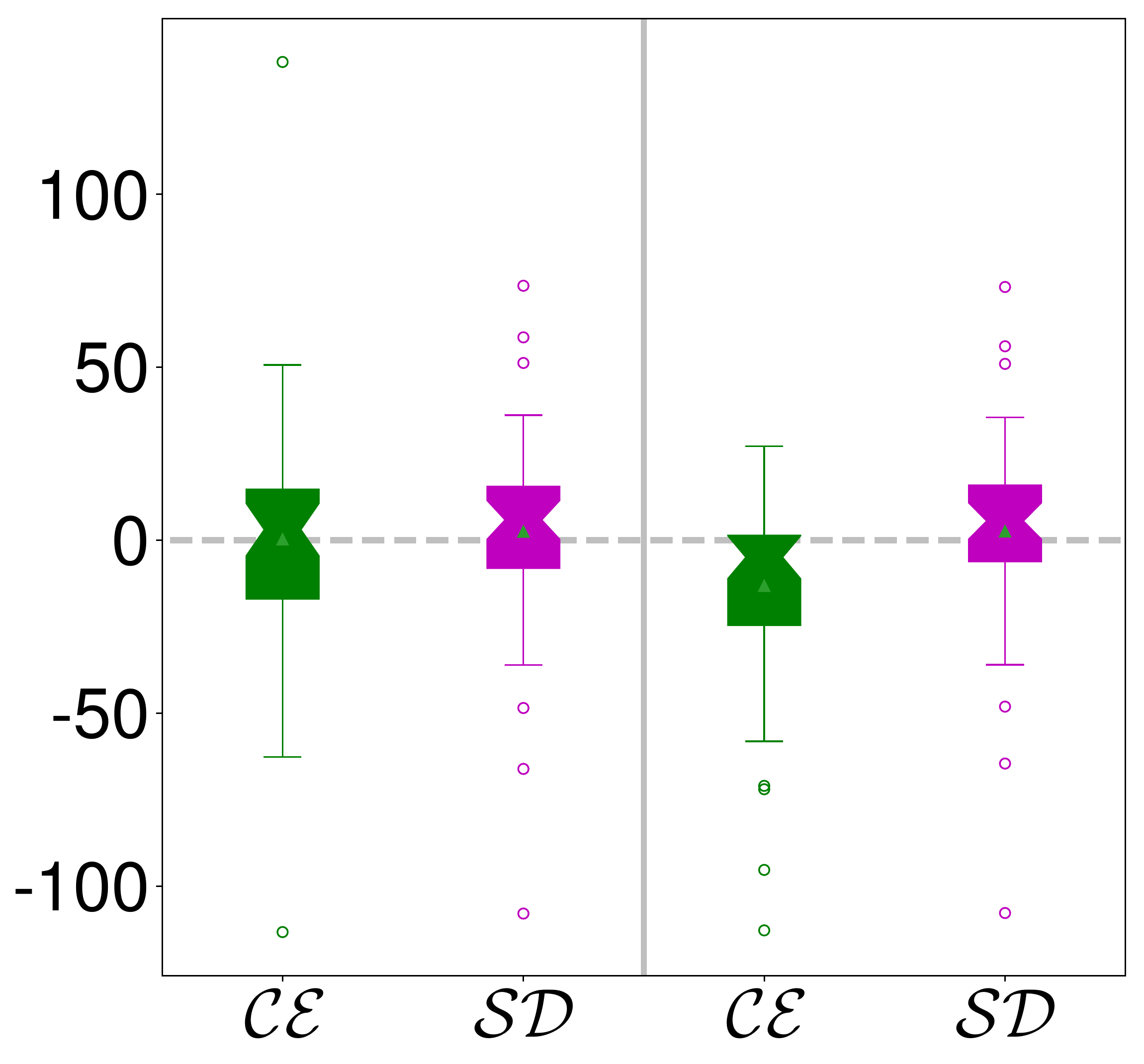}}
&\raisebox{-0.5\height}{\includegraphics[scale=\scale,scale=\myscale]{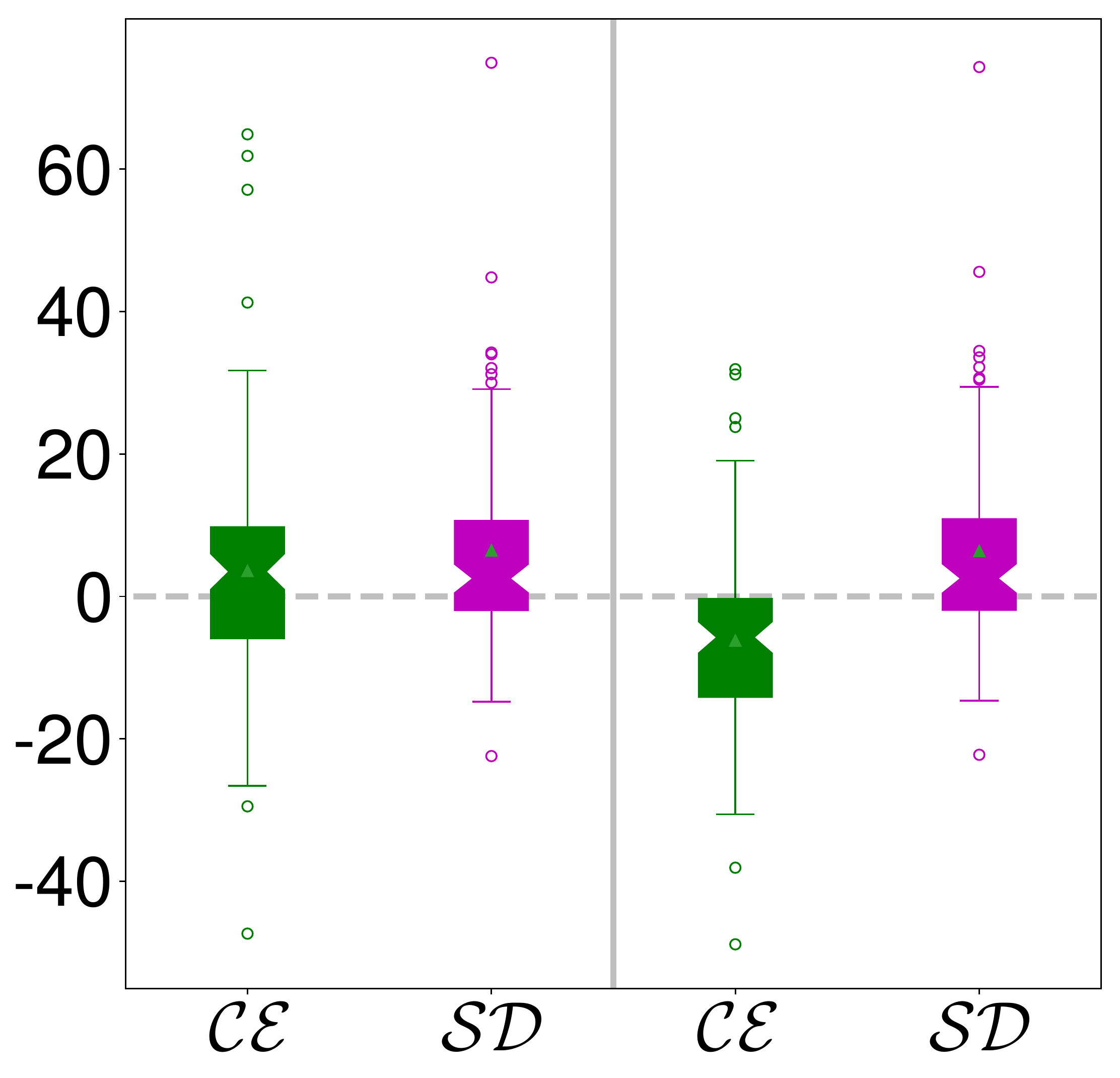}}
\end{tabular}
\caption[Boxplots of the volume errors.]{Boxplots of the volume errors $\Delta\mathcal{V}$ after $\mathcal{CE}$ (green) or $\mathcal{SD}$ (magenta) optimization for U-Net L trained on four medical tasks (from left to right: MO17, BR18, IS17 and IS18). The left side and right side in each boxplot display $\Delta\mathcal{V}(\tilde{y},l)$ and $\Delta\mathcal{V}(\tilde{l},l)$, respectively.}
\label{fig:boxplots}
\end{figure}

%% file: tables/table_other_measures.tex
\begin{table}[t]
\setlength{\tabcolsep}{3pt}
\centering
\caption[Results for other measures after $\mathcal{CE}$ or $\mathcal{SD}$ optimization.]{Results for other measures after $\mathcal{CE}$ or $\mathcal{SD}$ optimization for U-Net M and U-Net L trained on four medical tasks. Underlined values reflect a significantly better result. Each optimization objective indeed optimizes its corresponding loss and related target metric, respectively $\mathbb{E}[0/1]$ and $\mathbb{E}[\mathcal{D}]$. The relative volume bias $\mathbb{E}[\Delta\mathcal{V}/\mathcal{V}(l)]$, absolute volume bias $\mathbb{E}|\Delta\mathcal{V}|$ and relative absolute volume bias $\mathbb{E}[|\Delta\mathcal{V}|/\mathcal{V}(l)]$ are calculated for $\mathcal{CE}$ and $\mathcal{SD}$ with predictions $\tilde{y}$ or $\tilde{l}$, respectively.}
\label{tab:table_other_measures}
\begin{tabular}{ll|rr|rr}
\toprule
& \multicolumn{1}{r|}{Model $\rightarrow$} & \multicolumn{2}{c|}{U-Net M} & \multicolumn{2}{c}{U-Net L} \\
& \multicolumn{1}{r|}{Training loss $\rightarrow$} & \multicolumn{1}{c}{$\mathcal{CE}$} & \multicolumn{1}{c|}{$\mathcal{SD}$} & \multicolumn{1}{c}{$\mathcal{CE}$} & \multicolumn{1}{c}{$\mathcal{SD}$} \\
Dataset $\downarrow$ & Metric $\downarrow$ & & & & \\
\midrule
\multirow{7}{*}{MO17} & $\mathbb{E}[\mathcal{CE}]$ & \underline{0.024} & 0.103 & \underline{0.021} & 0.071 \\
     & $\mathbb{E}[0/1]$ & 0.991 & 0.992 & 0.993 & 0.993 \\
     & $1-\mathbb{E}[\mathcal{SD}]$ & 0.865 & \underline{0.931} & 0.879 & \underline{0.943} \\
     & $\mathbb{E}[\mathcal{D}]$ & 0.924 & \underline{0.932} & 0.942 & 0.943 \\
     & $\mathbb{E}[\Delta\mathcal{V}/\mathcal{V}(l)]$ & 0.062 & \underline{0.033} & 0.050 & \underline{0.026} \\
     & $\mathbb{E}|\Delta\mathcal{V}|$ ($10^3$ pixels) & 2.963 & \underline{2.249} & 2.608 & \underline{1.758} \\
     & $\mathbb{E}[|\Delta\mathcal{V}|/\mathcal{V}(l)]$ & 0.131 & \underline{0.099} & 0.114 & \underline{0.075} \\
\cline{1-6}
\multirow{7}{*}{BR18} & $\mathbb{E}[\mathcal{CE}]$ & \underline{0.012} & 0.027 & \underline{0.009} & 0.023 \\
     & $\mathbb{E}[0/1]$ & 0.996 & 0.997 & 0.997 & 0.998 \\
     & $1-\mathbb{E}[\mathcal{SD}]$ & 0.585 & \underline{0.820} & 0.685 & \underline{0.872} \\
     & $\mathbb{E}[\mathcal{D}]$ & 0.763 & \underline{0.826} & 0.840 & \underline{0.879} \\
     & $\mathbb{E}[\Delta\mathcal{V}/\mathcal{V}(l)]$ & 0.093 & \underline{0.014} & 0.140 & \underline{0.004} \\
     & $\mathbb{E}|\Delta\mathcal{V}|$ (ml) & 22.145 & \underline{15.787} & 19.667 & \underline{12.095} \\
     & $\mathbb{E}[|\Delta\mathcal{V}|/\mathcal{V}(l)]$ & 0.301 & \underline{0.195} & 0.286 & \underline{0.135} \\
\cline{1-6}
\multirow{7}{*}{IS17} & $\mathbb{E}[\mathcal{CE}]$ & \underline{0.014} & 0.066 & \underline{0.014} & 0.043 \\
     & $\mathbb{E}[0/1]$ & \underline{0.996} & 0.993 & 0.995 & 0.995 \\
     & $1-\mathbb{E}[\mathcal{SD}]$ & 0.188 & \underline{0.340} & 0.230 & \underline{0.374} \\
     & $\mathbb{E}[\mathcal{D}]$ & 0.177 & \underline{0.343} & 0.279 & \underline{0.383} \\
     & $\mathbb{E}[\Delta\mathcal{V}/\mathcal{V}(l)]$ & \underline{5.803} & 18.731 & \underline{4.742} & 11.164 \\
     & $\mathbb{E}|\Delta\mathcal{V}|$ (ml) & 23.228 & 36.188 & 26.119 & 21.669 \\
     & $\mathbb{E}[|\Delta\mathcal{V}|/\mathcal{V}(l)]$ & \underline{6.264} & 19.124 & \underline{5.108} & 11.435 \\
\cline{1-6}
\multirow{7}{*}{IS18} & $\mathbb{E}[\mathcal{CE}]$ & \underline{0.029} & 0.128 & \underline{0.031} & 0.145 \\
     & $\mathbb{E}[0/1]$ & \underline{0.989} & 0.987 & 0.989 & 0.988 \\
     & $1-\mathbb{E}[\mathcal{SD}]$ & 0.362 & \underline{0.518} & 0.344 & \underline{0.538} \\
     & $\mathbb{E}[\mathcal{D}]$ & 0.454 & \underline{0.527} & 0.409 & \underline{0.549} \\
     & $\mathbb{E}[\Delta\mathcal{V}/\mathcal{V}(l)]$ & \underline{1.069} & 1.578 & 1.857 & \underline{1.753} \\
     & $\mathbb{E}|\Delta\mathcal{V}|$ (ml) & 11.453 & 11.935 & 12.421 & \underline{10.725} \\
     & $\mathbb{E}[|\Delta\mathcal{V}|/\mathcal{V}(l)]$ & \underline{1.349} & 1.870 & 2.042 & \underline{1.894} \\
\bottomrule
\end{tabular}
\end{table}

%% file: tables/table_CE_pretrain.tex
\begin{table}[t]
\setlength{\tabcolsep}{3pt}
\centering
\caption[Effect of $\mathcal{CE}$ pre-training on the volume bias after $\mathcal{CE}$ or $\mathcal{SD}$ optimization.]{Effect of $\mathcal{CE}$ pre-training on the volume bias $\mathbb{E}[\Delta\mathcal{V}]$ after $\mathcal{CE}$ or $\mathcal{SD}$ optimization for U-Net M and U-Net L trained on four medical tasks, calculated with predictions $\tilde{y}$ or $\tilde{l}$, respectively. Values closer to zero are better and bold or italic are values significantly greater or smaller than zero.}
\label{tab:table_CE_pre-train}
\begin{tabular}{ll|rr|rr}
\toprule
& \multicolumn{1}{r|}{Model $\rightarrow$} & \multicolumn{2}{c|}{U-Net M} & \multicolumn{2}{c}{U-Net L} \\
& \multicolumn{1}{r|}{Training loss $\rightarrow$} & \multicolumn{1}{c}{$\mathcal{CE}$} & \multicolumn{1}{c|}{$\mathcal{SD}$} & \multicolumn{1}{c}{$\mathcal{CE}$} & \multicolumn{1}{c}{$\mathcal{SD}$} \\
Dataset $\downarrow$ & Metric $\downarrow$ & & \\
\midrule
MO17 & $\mathbb{E}[\Delta\mathcal{V}]$ ($10^3$ pixels) & -0.22 & 0.05 & \textit{-0.31} & -0.02 \\
BR18 & $\mathbb{E}[\Delta\mathcal{V}]$ (ml) & \textit{-8.69} & -2.13 & \textit{-5.13} & \textit{-5.16} \\
IS17 & $\mathbb{E}[\Delta\mathcal{V}]$ (ml) & 1.84 & \textbf{20.84} & 1.15 & 15.65 \\
IS18 & $\mathbb{E}[\Delta\mathcal{V}]$ (ml) & -0.06 & \textbf{3.24} & -0.05 & \textbf{3.24} \\
\bottomrule
\end{tabular}
\end{table}

%% file: figures/qualitative_inspection_ce_dice.tex
\begin{figure}[t!]
\newcommand\myscale{0.8}
\setlength\tabcolsep{0pt}
\centering
\includegraphics[scale=\scale,scale=\myscale]{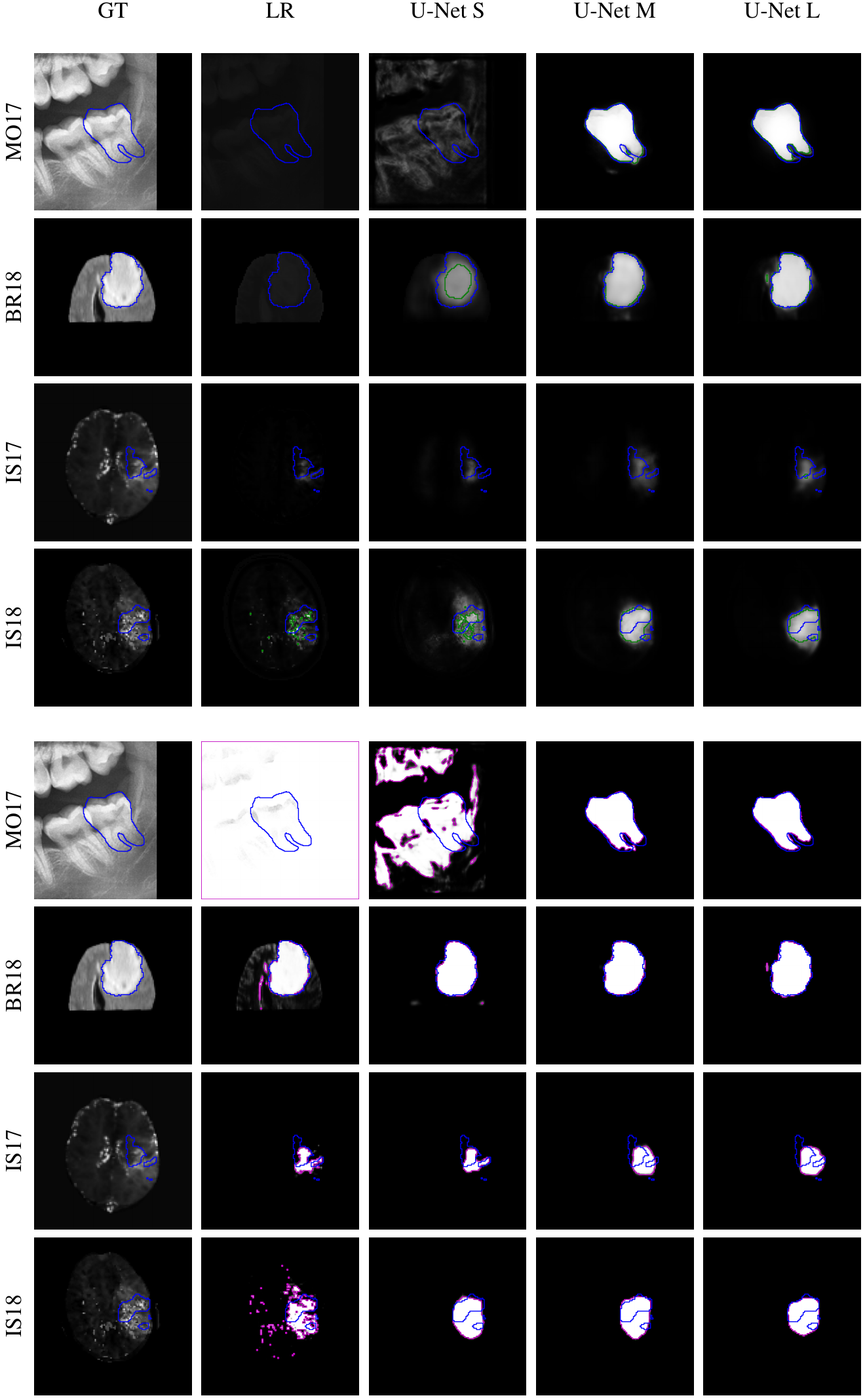}
\caption[Qualitative inspection of the effect of inherent uncertainty versus loss function.]{Qualitative inspection of the input ($x$) and predicted segmentations ($\tilde{y}$) after $\mathcal{CE}$ (TOP; green) or $\mathcal{SD}$ (BOTTOM; magenta) optimization for the four models (columns) trained on four medical tasks (rows). The true delineation is depicted in blue and the predicted delineation after thresholding at 0.5 ($\tilde{l}$) in red. Slices shown are representative examples and were selected based on the heuristic having a median $\Delta\mathcal{V}$ across the slices (containing at least one true foreground voxel) in the volume that had a median $\Delta\mathcal{V}$ across the dataset based on the U-Net L model after $\mathcal{CE}$ optimization. For $\mathcal{CE}$, less inherent uncertainty condenses the predicted soft segmentation maps. For $\mathcal{SD}$, we can see binary-like maps and over-segmentation.}
\label{fig:qualitative_inspection_ce_dice}
\end{figure}

%% file: figures/scale_specific_study_and_recalibration.tex
\begin{figure}[b!]
\newcommand\myscale{0.17}
\setlength\tabcolsep{0pt}
\centering
\begin{tabular}{ccccc}
    & \quad MO17 & \quad BR18 & \quad IS17 & \quad IS18\\
    \\
    \rotatebox[origin=c]{90}{\textcolor{green}{$\mathcal{V}(\tilde{y})$} or \textcolor{magenta}{$\mathcal{V}(\tilde{l})$}}
    &\raisebox{-0.5\height}{\includegraphics[scale=\scale,scale=\myscale]{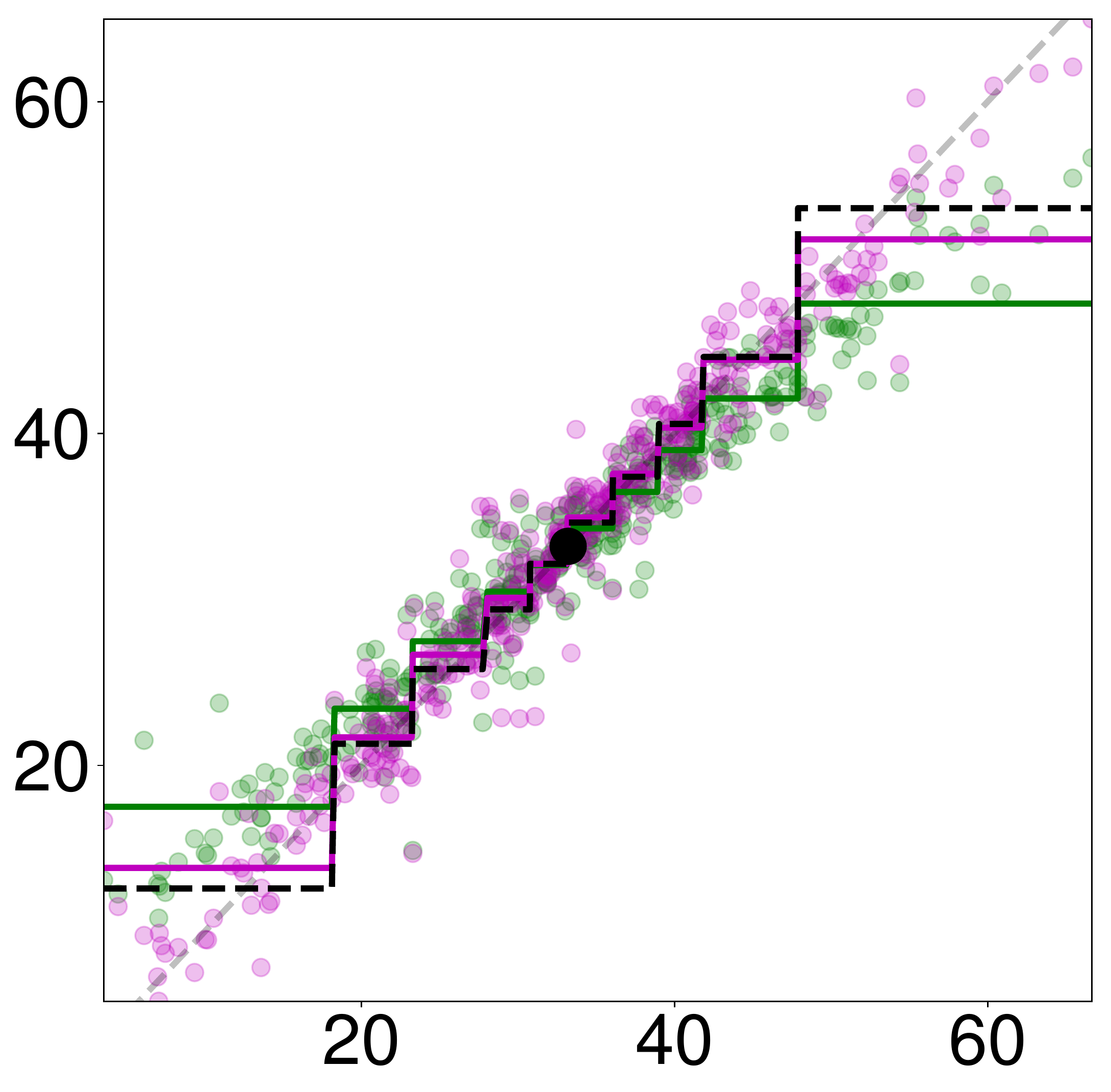}}
    &\raisebox{-0.5\height}{\includegraphics[scale=\scale,scale=\myscale]{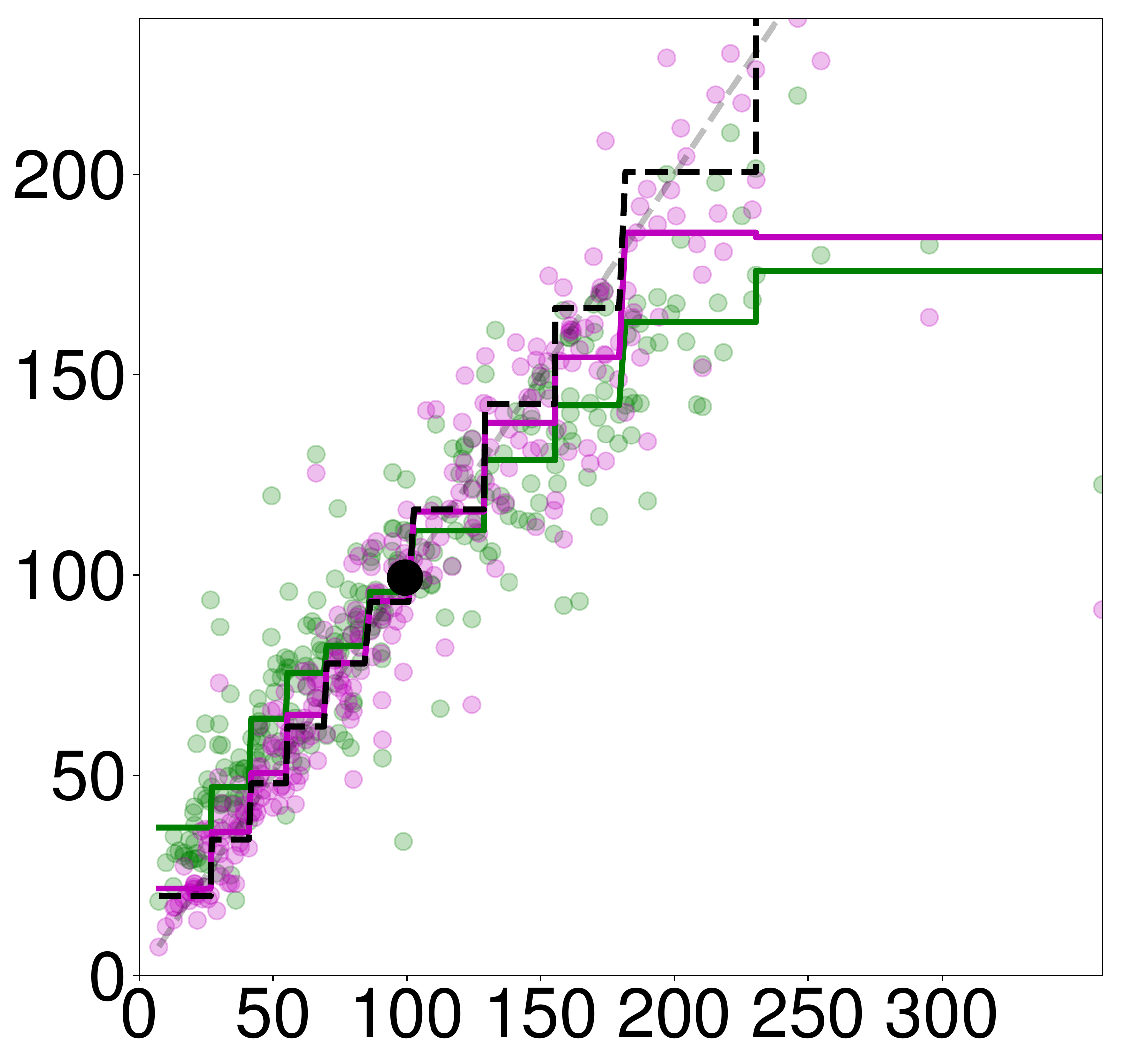}}
    &\raisebox{-0.5\height}{\includegraphics[scale=\scale,scale=\myscale]{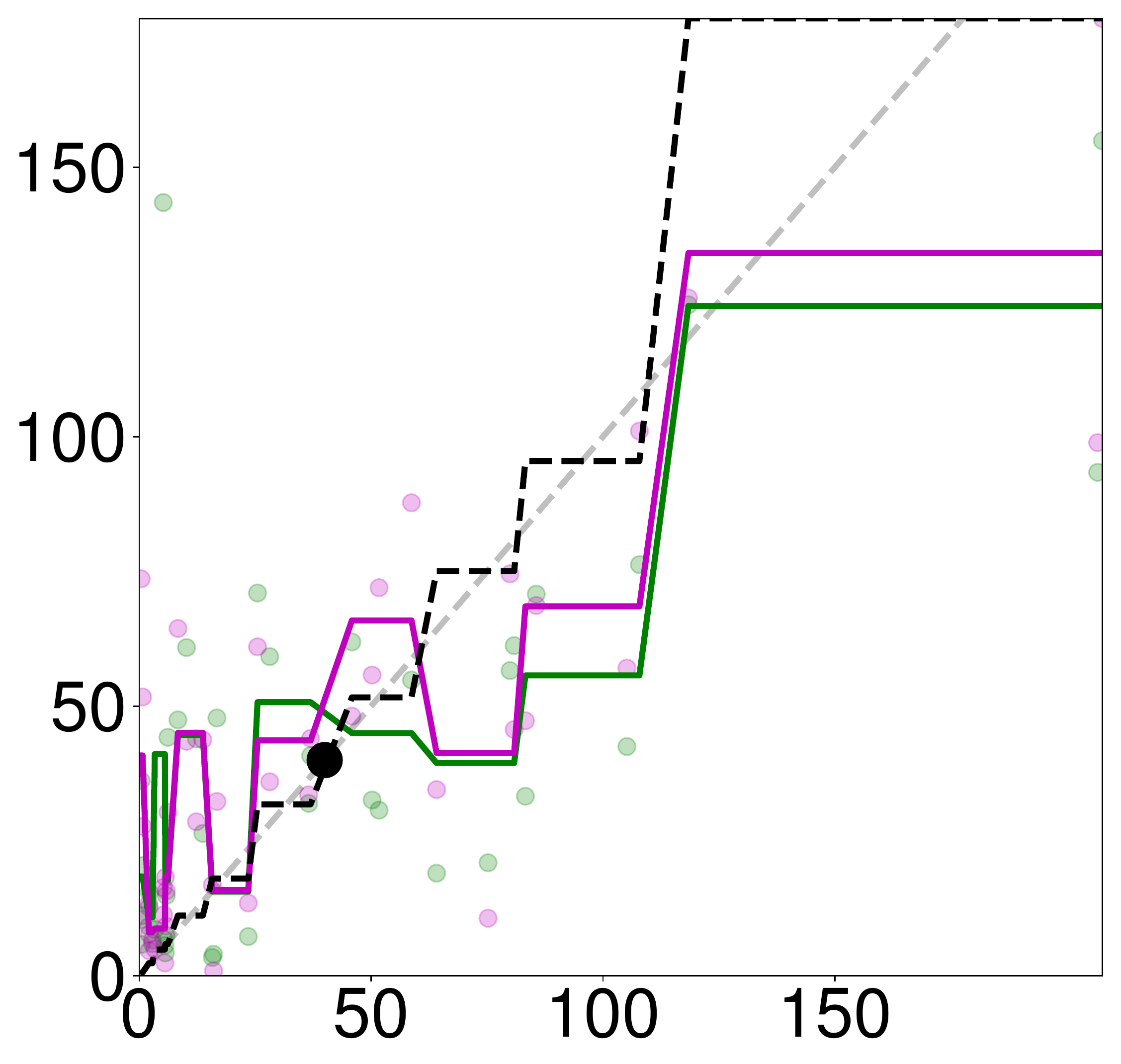}}
    &\raisebox{-0.5\height}{\includegraphics[scale=\scale,scale=\myscale]{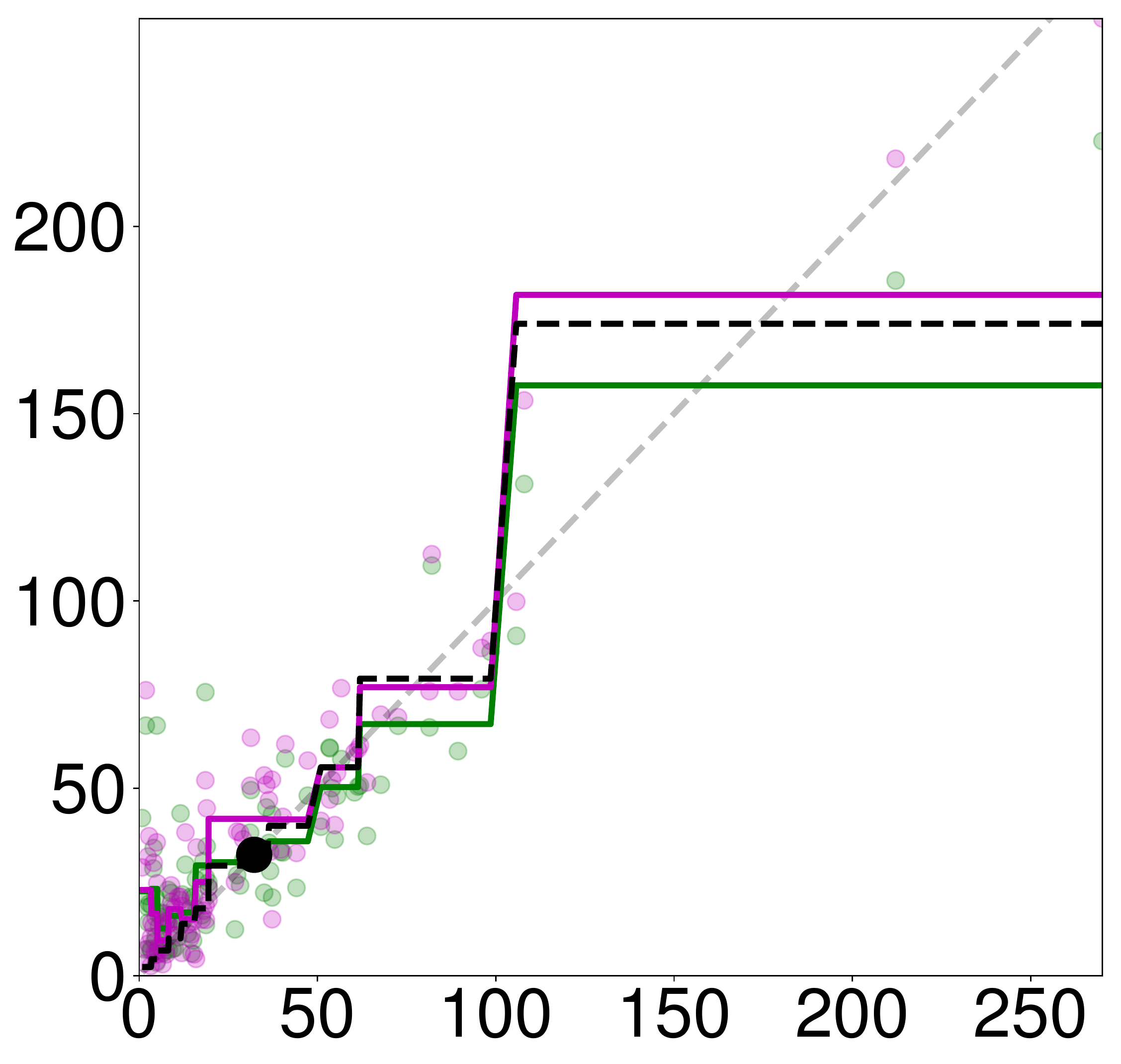}}\\
    \\
    \rotatebox[origin=c]{90}{\textcolor{green}{$\mathcal{V}(\tilde{y})$} or \textcolor{magenta}{$\mathcal{V}(\tilde{l})$}}
    &\raisebox{-0.5\height}{\includegraphics[scale=\scale,scale=\myscale]{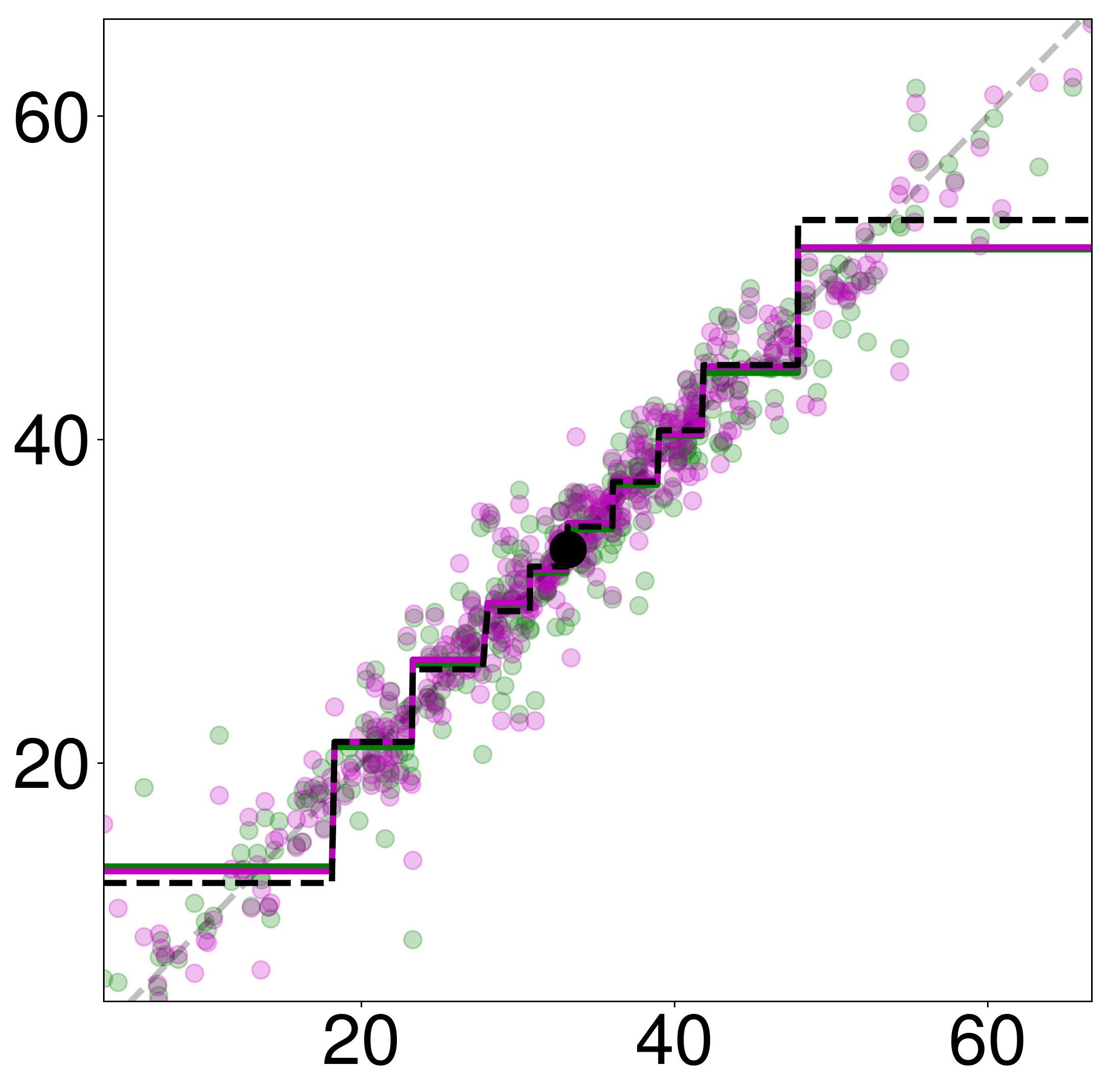}}
    &\raisebox{-0.5\height}{\includegraphics[scale=\scale,scale=\myscale]{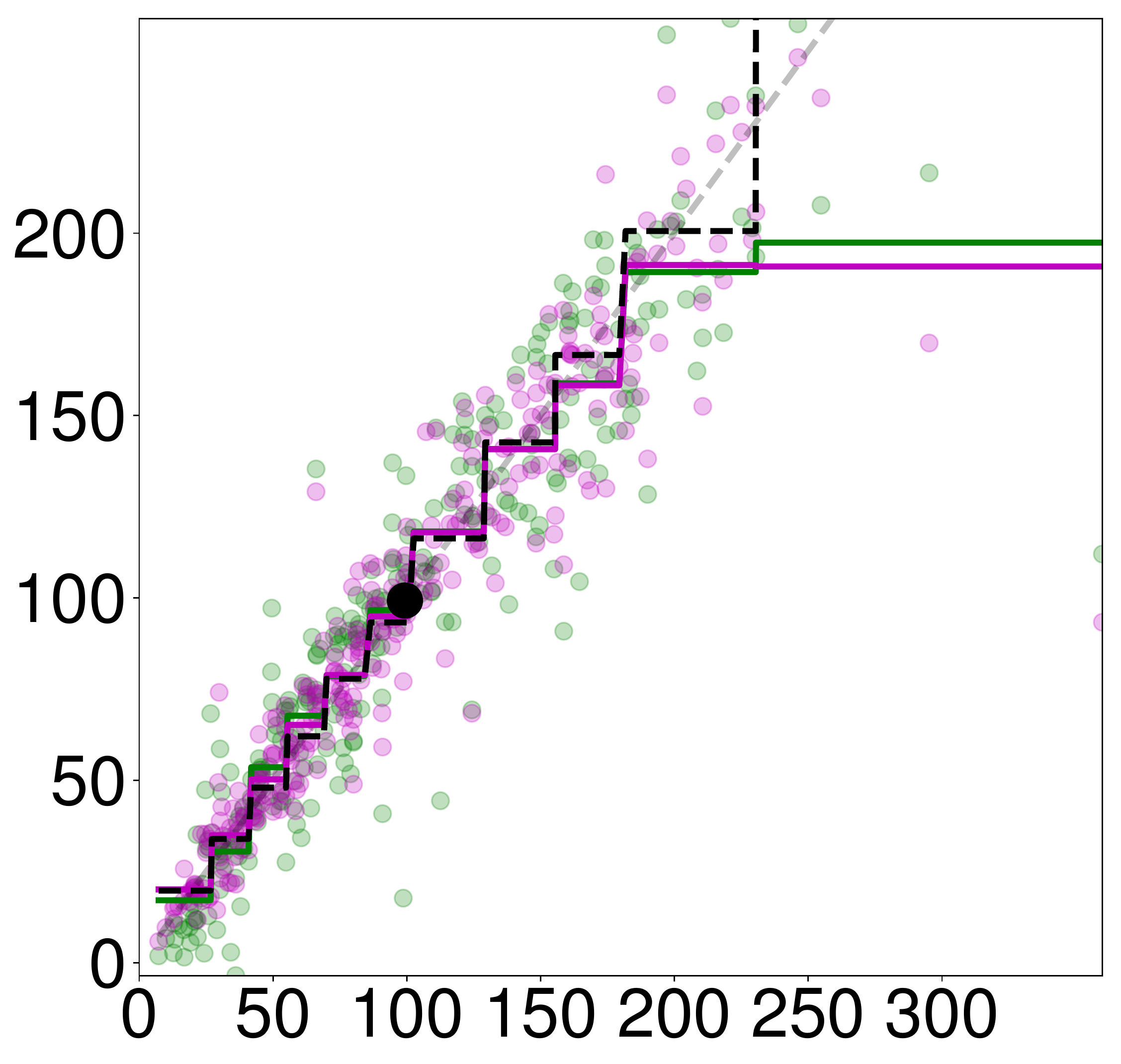}}
    &\raisebox{-0.5\height}{\includegraphics[scale=\scale,scale=\myscale]{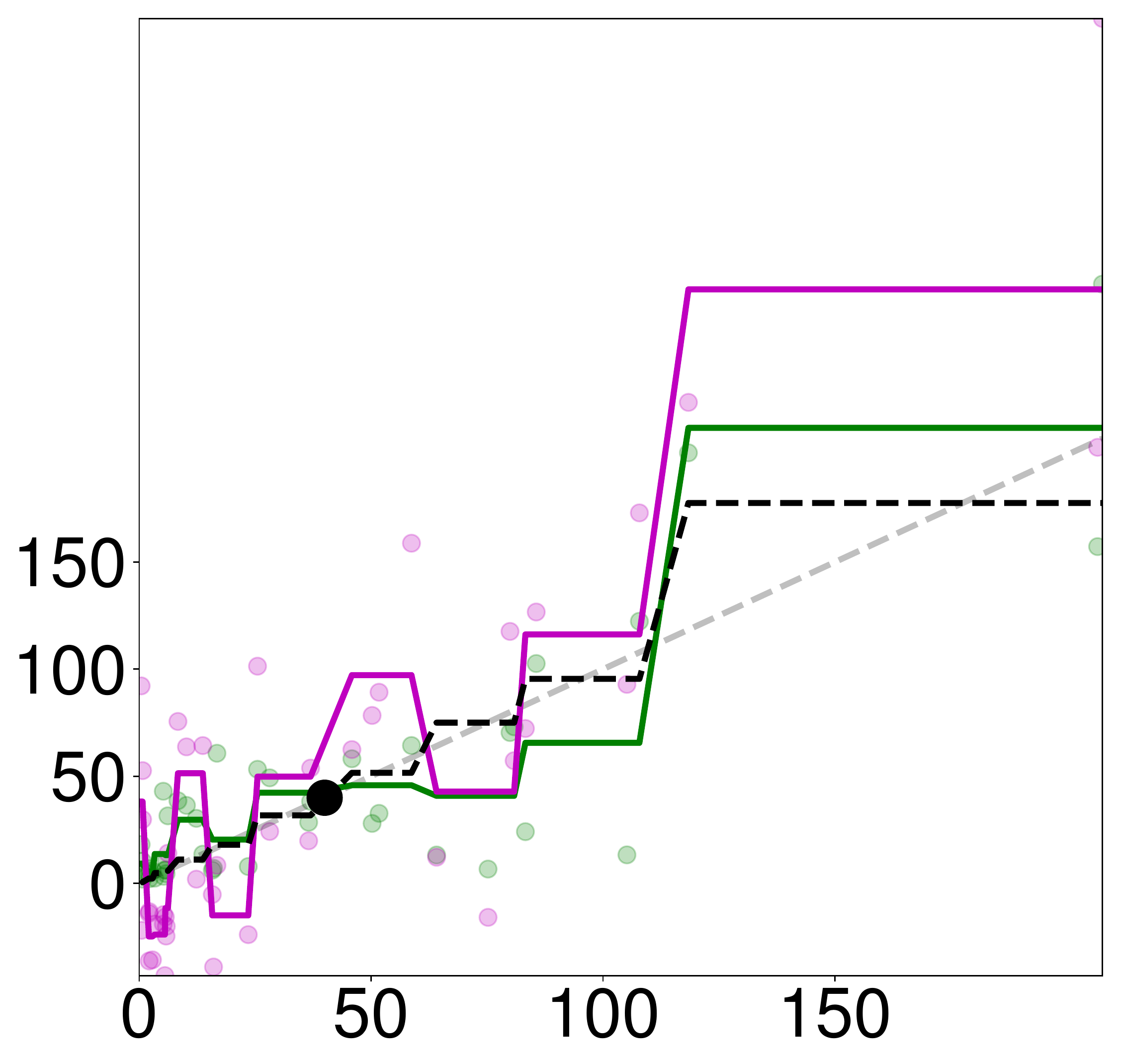}}
    &\raisebox{-0.5\height}{\includegraphics[scale=\scale,scale=\myscale]{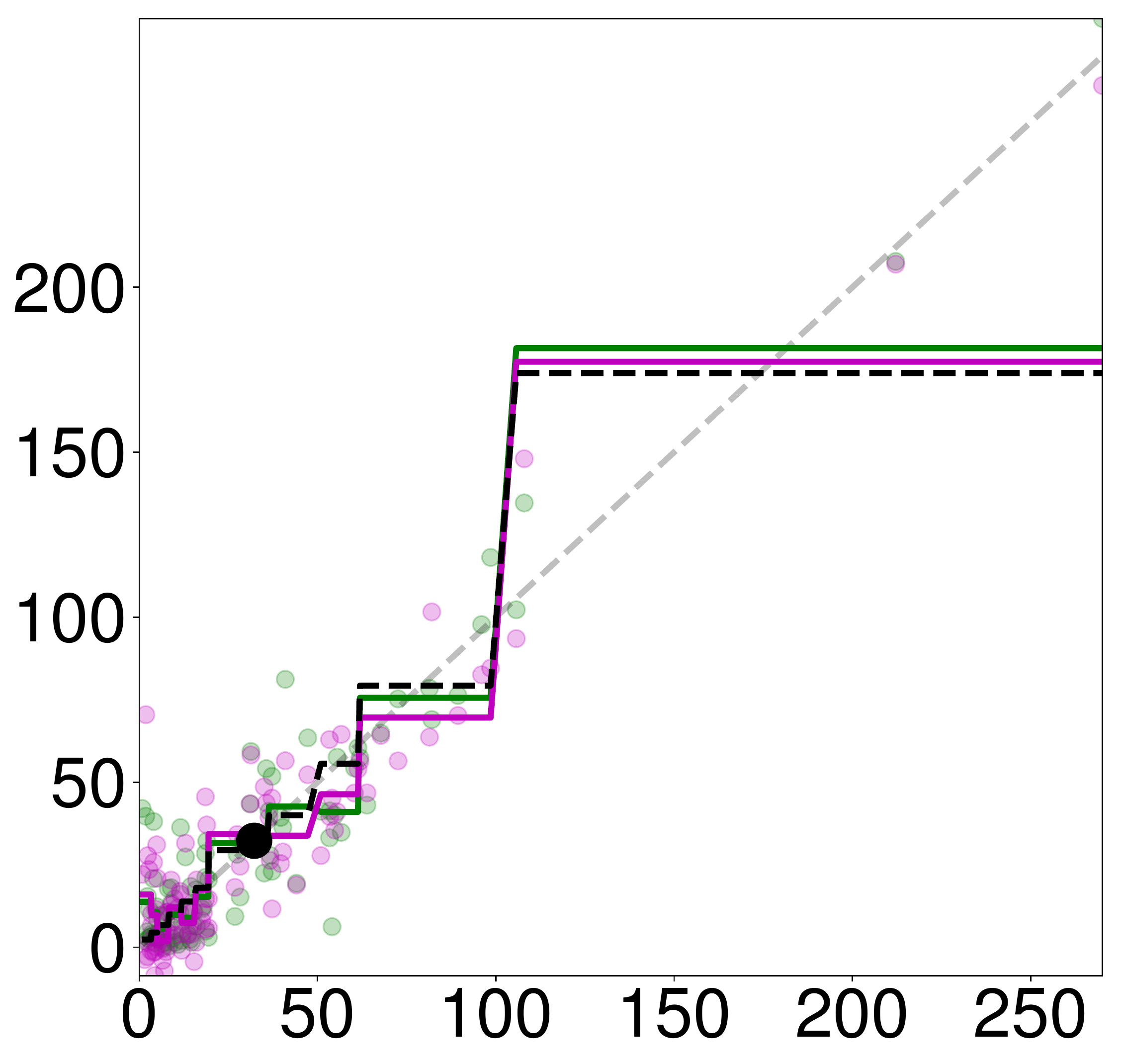}}\\
    & \quad $\mathcal{V}(l)$ & \quad $\mathcal{V}(l)$ & \quad $\mathcal{V}(l)$ & \quad $\mathcal{V}(l)$
\end{tabular}
\caption[Volume-specific analysis and recalibration.]{Volume-specific analysis of the predicted volumes $\mathcal{V}(\tilde{y})$ or $\mathcal{V}(\tilde{l})$ after $\mathcal{CE}$ (green) or $\mathcal{SD}$ (magenta) optimization, respectively, with U-Net L on four medical segmentation tasks (from left to right: MO17, BR18, IS17 and IS18) before (TOP) and after (BOTTOM) re-calibration. Small dots represent the individual data points, while the lines are averages across each 10th percentile. The large black dot depicts the average true volume. Both $\mathcal{CE}$ and $\mathcal{SD}$ optimization suffer from volume-specific errors, with a turning point around the average true volume. The predictions after $\mathcal{SD}$ optimization seem to lie more closely to the desired unity line (dashed line). The re-calibration as proposed under Sect. \ref{sec:re-calibration} effectively reduces the volume-specific errors. In this case, re-calibration for each fold was performed using the volume-specific errors present in the training set.}
\label{fig:scale_specific_study_and_recalibration}
\end{figure}